\PassOptionsToPackage{unicode}{hyperref}
\PassOptionsToPackage{hyphens}{url}
\PassOptionsToPackage{dvipsnames,svgnames,x11names}{xcolor}
\documentclass[
  11pt,
  letterpaper,
]{article}
\usepackage{xcolor}
\usepackage[margin=1in]{geometry}
\usepackage{amsmath,amssymb}
\usepackage{iftex}
\ifPDFTeX
  \usepackage[T1]{fontenc}
  \usepackage{textcomp} % provide euro and other symbols
\else % if luatex or xetex
  \usepackage{unicode-math} % this also loads fontspec
  \defaultfontfeatures{Scale=MatchLowercase}
  \defaultfontfeatures[\rmfamily]{Ligatures=TeX,Scale=1}
\fi
\usepackage{lmodern}
\ifPDFTeX\else
\fi
\IfFileExists{upquote.sty}{\usepackage{upquote}}{}
\IfFileExists{microtype.sty}{% use microtype if available
  \usepackage[]{microtype}
  \UseMicrotypeSet[protrusion]{basicmath} % disable protrusion for tt fonts
}{}
\makeatletter
\@ifundefined{KOMAClassName}{% if non-KOMA class
  \IfFileExists{parskip.sty}{%
    \usepackage{parskip}
  }{% else
    \setlength{\parindent}{0pt}
    \setlength{\parskip}{6pt plus 2pt minus 1pt}}
}{% if KOMA class
  \KOMAoptions{parskip=half}}
\makeatother
\makeatletter
\ifx\paragraph\undefined\else
  \let\oldparagraph\paragraph
  \renewcommand{\paragraph}{
    \@ifstar
      \xxxParagraphStar
      \xxxParagraphNoStar
  }
  \newcommand{\xxxParagraphStar}[1]{\oldparagraph*{#1}\mbox{}}
  \newcommand{\xxxParagraphNoStar}[1]{\oldparagraph{#1}\mbox{}}
\fi
\ifx\subparagraph\undefined\else
  \let\oldsubparagraph\subparagraph
  \renewcommand{\subparagraph}{
    \@ifstar
      \xxxSubParagraphStar
      \xxxSubParagraphNoStar
  }
  \newcommand{\xxxSubParagraphStar}[1]{\oldsubparagraph*{#1}\mbox{}}
  \newcommand{\xxxSubParagraphNoStar}[1]{\oldsubparagraph{#1}\mbox{}}
\fi
\makeatother

\usepackage{longtable,booktabs,array}
\usepackage{calc} % for calculating minipage widths
\usepackage{etoolbox}
\makeatletter
\patchcmd\longtable{\par}{\if@noskipsec\mbox{}\fi\par}{}{}
\makeatother
\IfFileExists{footnotehyper.sty}{\usepackage{footnotehyper}}{\usepackage{footnote}}
\makesavenoteenv{longtable}
\usepackage{graphicx}
\makeatletter
\newsavebox\pandoc@box
\newcommand*\pandocbounded[1]{% scales image to fit in text height/width
  \sbox\pandoc@box{#1}%
  \Gscale@div\@tempa{\textheight}{\dimexpr\ht\pandoc@box+\dp\pandoc@box\relax}%
  \Gscale@div\@tempb{\linewidth}{\wd\pandoc@box}%
  \ifdim\@tempb\p@<\@tempa\p@\let\@tempa\@tempb\fi% select the smaller of both
  \ifdim\@tempa\p@<\p@\scalebox{\@tempa}{\usebox\pandoc@box}%
  \else\usebox{\pandoc@box}%
  \fi%
}
\def\fps@figure{htbp}
\makeatother

\NewDocumentCommand\citeproctext{}{}

\makeatletter
 \let\@cite@ofmt\@firstofone
 \def\@biblabel#1{}
 \def\@cite#1#2{{#1\if@tempswa , #2\fi}}
\makeatother
\newlength{\cslhangindent}
\newlength{\csllabelwidth}
\newenvironment{CSLReferences}[2] % #1 hanging-indent, #2 entry-spacing
 {\begin{list}{}{%
  \setlength{\itemindent}{0pt}
  \setlength{\leftmargin}{0pt}
  \setlength{\parsep}{0pt}
  \ifodd #1
   \setlength{\leftmargin}{\cslhangindent}
   \setlength{\itemindent}{-1\cslhangindent}
  \fi
  \setlength{\itemsep}{#2\baselineskip}}}
 {\end{list}}
\usepackage{calc}

\DeclareUnicodeCharacter{2212}{\ensuremath{-}}
\DeclareUnicodeCharacter{2013}{--}
\DeclareUnicodeCharacter{2014}{---}
\DeclareUnicodeCharacter{2009}{\,}
\DeclareUnicodeCharacter{00A0}{~}
\DeclareUnicodeCharacter{00D7}{\ensuremath{\times}}
\DeclareUnicodeCharacter{2264}{\ensuremath{\leq}}
\DeclareUnicodeCharacter{2265}{\ensuremath{\geq}}
\DeclareUnicodeCharacter{2260}{\ensuremath{\neq}}
\DeclareUnicodeCharacter{2248}{\ensuremath{\approx}}
\DeclareUnicodeCharacter{03B1}{\ensuremath{\alpha}}
\DeclareUnicodeCharacter{03B2}{\ensuremath{\beta}}
\DeclareUnicodeCharacter{03C3}{\ensuremath{\sigma}}
\DeclareUnicodeCharacter{03C7}{\ensuremath{\chi}}
\DeclareUnicodeCharacter{2032}{\ensuremath{'}}
\DeclareUnicodeCharacter{2713}{\checkmark}

\usepackage{caption}
\usepackage{multirow}
\AtBeginEnvironment{apptbl}{\footnotesize}
\makeatletter
\newcommand\table@defaultsize{\footnotesize}
\let\table@size\table@defaultsize
\newcommand\nexttablesize[1]{\gdef\table@size{#1}}
\AtBeginEnvironment{longtable}{\table@size\global\let\table@size\table@defaultsize}
\makeatother

\makeatletter
\let\floatnote@text\@empty
\newcommand\floatnote[1]{\gdef\floatnote@text{#1}}
\newcommand\fs@plainnote{\fs@plain
  \def\@fs@post{\ifx\floatnote@text\@empty\else
    \vspace\abovecaptionskip\noindent\floatnote@text\par
    \global\let\floatnote@text\@empty\fi}}
\makeatother

\AtBeginDocument{%
  \floatstyle{plainnote}\restylefloat{appfig}%
  \floatplacement{appfig}{H}\floatplacement{apptbl}{H}}

\AtBeginDocument{}
\makeatletter
\renewcommand\@maketitle{%
  \newpage\null\vskip 0.5em
  \begin{center}
    {\Large\bfseries \@title\par}
    \vskip 1.5em
    {\normalsize \@author\par}
  \end{center}
  \par\vskip 1em}
\newcommand\aclauthor[3]{% name, affiliation lines, email
  \begin{tabular}[t]{@{}c@{}}
    \textbf{#1}\\[1pt]
    #2\\[1pt]
    {\small\texttt{#3}}
  \end{tabular}}
\makeatother

\AtBeginDocument{%
  \author{%
    \aclauthor{Davood Wadi}
      {Desautels Faculty of Management\\ McGill University\\ Montreal, Quebec, Canada}
      {davood.wadi@hec.ca}%
    \hspace{3em}%
    \aclauthor{Yu Ma}
      {Desautels Faculty of Management\\ McGill University\\ Montreal, Quebec, Canada}
      {yu.ma@mcgill.ca}%
  }%
}
\makeatletter
\@ifpackageloaded{float}{}{\usepackage{float}}
\floatstyle{plain}
\@ifundefined{c@chapter}{\newfloat{apptbl}{h}{loapptbl}}{\newfloat{apptbl}{h}{loapptbl}[chapter]}
\floatname{apptbl}{Table W}
\floatstyle{plaintop}
\restylefloat{apptbl}
\newcommand*\quartoapptblref[1]{Table \hyperref[#1]{W\ref{#1}}}
\@ifpackageloaded{caption}{}{\usepackage{caption}}
\DeclareCaptionLabelFormat{quartoapptblreflabelformat}{#1#2}
\makeatother
\makeatletter
\@ifpackageloaded{float}{}{\usepackage{float}}
\floatstyle{plain}
\@ifundefined{c@chapter}{\newfloat{appfig}{h}{loappfig}}{\newfloat{appfig}{h}{loappfig}[chapter]}
\floatname{appfig}{Figure W}
\newcommand*\quartoappfigref[1]{Figure \hyperref[#1]{W\ref{#1}}}
\@ifpackageloaded{caption}{}{\usepackage{caption}}
\DeclareCaptionLabelFormat{quartoappfigreflabelformat}{#1#2}
\makeatother
\makeatletter
\@ifpackageloaded{float}{}{\usepackage{float}}
\floatstyle{plain}
\@ifundefined{c@chapter}{\newfloat{appenv}{h}{loapp}}{\newfloat{appenv}{h}{loapp}[chapter]}
\floatname{appenv}{Web Appendix}
\floatstyle{plaintop}
\restylefloat{appenv}
\newcommand*\quartoappref[1]{Web \hyperref[#1]{Appendix\ref{#1}}}
\@ifpackageloaded{caption}{}{\usepackage{caption}}
\DeclareCaptionLabelFormat{quartoappreflabelformat}{#1#2}
\makeatother
\makeatletter
\@ifpackageloaded{caption}{}{\usepackage{caption}}
\AtBeginDocument{%
\ifdefined\contentsname
  \renewcommand*\contentsname{Table of contents}
\else
  \newcommand\contentsname{Table of contents}
\fi
\ifdefined\listfigurename
  \renewcommand*\listfigurename{List of Figures}
\else
  \newcommand\listfigurename{List of Figures}
\fi
\ifdefined\listtablename
  \renewcommand*\listtablename{List of Tables}
\else
  \newcommand\listtablename{List of Tables}
\fi
\ifdefined\figurename
  \renewcommand*\figurename{Figure}
\else
  \newcommand\figurename{Figure}
\fi
\ifdefined\tablename
  \renewcommand*\tablename{Table}
\else
  \newcommand\tablename{Table}
\fi
}
\@ifpackageloaded{float}{}{\usepackage{float}}
\floatstyle{ruled}
\@ifundefined{c@chapter}{\newfloat{codelisting}{h}{lop}}{\newfloat{codelisting}{h}{lop}[chapter]}
\floatname{codelisting}{Listing}

\makeatother
\makeatletter
\@ifpackageloaded{caption}{}{\usepackage{caption}}
\@ifpackageloaded{subcaption}{}{\usepackage{subcaption}}
\makeatother
\usepackage{bookmark}
\IfFileExists{xurl.sty}{\usepackage{xurl}}{} % add URL line breaks if available
\makeatletter
\@ifundefined{xmpquote}{}{}
\makeatother
\hypersetup{
  pdftitle={Training decision models for simulating human respondents},
  pdfauthor={Davood Wadi; Yu Ma},
  colorlinks=true,
  linkcolor={blue},
  filecolor={Maroon},
  citecolor={Blue},
  urlcolor={Blue},
  pdfcreator={LaTeX via pandoc}}

\title{Training decision models for simulating human respondents}
\author{Davood Wadi \and Yu Ma}
\date{}
\begin{document}
\maketitle

\textbf{Abstract.}\\
Researchers increasingly use large language models (LLMs) as synthetic
respondents in market research, conjoint analysis, and randomized
experiments. However, synthetic choice data from off-the-shelf
generative LLMs bias downstream econometric inference. Neither
proprietary models nor higher reasoning effort reliably reduces this
bias. We investigate decision models (System One architectures) for
silicon sampling. Using choice training, we update KEV model weights on
historical respondent-level choice microdata by conditional maximum
likelihood, before inference and without target-study data. Across three
model scales, training on two discrete choice experiments lowers
treatment-effect error by 22\% to 89\% on four held-out studies from new
domains (vaccines, nutrition labels, consumer rights, and AI) and new
response formats. A small choice-trained model outperforms a much larger
off-the-shelf model on three of the four studies. Although training on
single studies lowers cross-entropy, it can exacerbate treatment-effect
error. Choice training thus complements post-inference calibration by
providing a stronger starting point for estimating human preferences.


\section{Introduction}\label{sec-intro}

\begin{figure}

\centering{

\includegraphics[width=0.9\linewidth,height=\textheight,keepaspectratio]{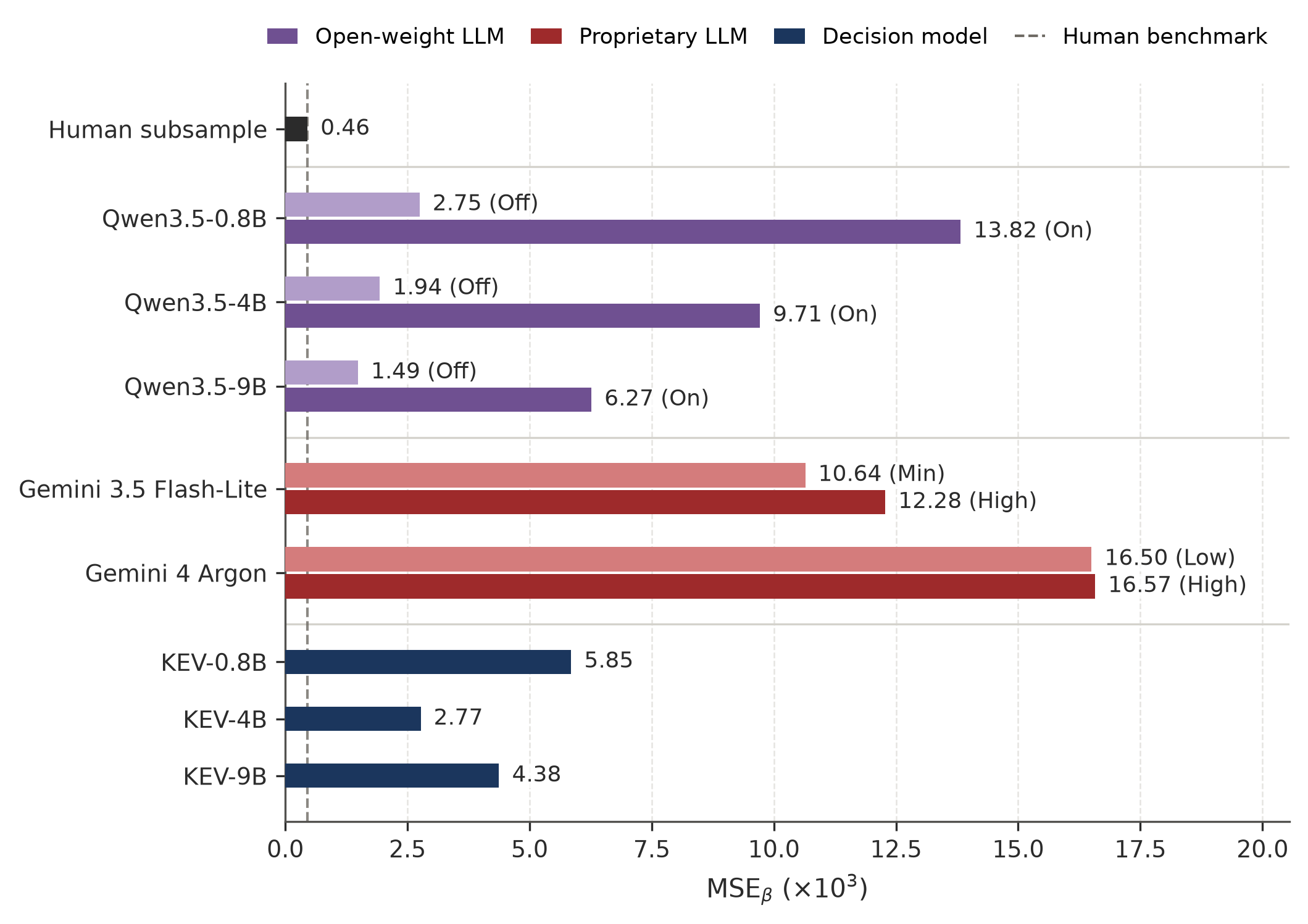}

}

\caption{\label{fig-intro-off-the-shelf}OLS treatment-effect error
across off-the-shelf generative LLMs and decision models}

\vspace{\abovecaptionskip}

\noindent \emph{Note.} Bars report OLS
\(\mathrm{MSE}_\beta \times 10^3\) (Equation~\ref{eq-eval-mse}) across
11 vaccine attribute levels on a 200-respondent Vaccines conjoint
subsample (\(n = 1,000\) choice tasks; Wang et al. 2026). Errors are
measured relative to full-sample human estimates (\(N = 1,971\)
respondents; 9,855 choice tasks).

\end{figure}

Using foundation models as synthetic respondents has attracted growing
interest in marketing and behavioral research. In marketing, a primary
goal is downstream econometric inference, such as estimating attribute
slopes in conjoint analysis or recovering causal treatment effects in
randomized experiments. Most studies on synthetic respondents generate
synthetic choice data with off-the-shelf generative large language
models (LLMs), yet substituting these outputs for human responses
severely biases downstream analyses (Goli and Singh 2024, Leng et al.
2026, Wang et al. 2026). A likely source of this bias is how LLMs are
trained. During training, LLMs are rewarded for producing a single
correct answer on single-label objective benchmarks (e.g., mathematics,
factual questions) and for complying with safety rules. In these tasks,
any probability placed on other answers is considered error to be
minimized. In contrast, simulating human choices requires a probability
distribution over all plausible responses, because respondents with the
same observed characteristics can differ in their preferences.
Therefore, off-the-shelf LLMs tend to collapse their response
distribution onto a single answer and, because of their safety training,
sometimes refuse to answer. For instance, on a vaccine conjoint
subsample (Kreps et al. 2020), off-the-shelf models incur large ordinary
least squares (OLS) treatment-effect error (\(\mathrm{MSE}_\beta\))
relative to human respondents (Wang et al. 2026). Contrary to common
expectation, using proprietary models or increasing reasoning effort can
exacerbate this error (Figure~\ref{fig-intro-off-the-shelf}).

Moreover, generative LLMs are slow and expensive because generating text
tokens requires costly Monte Carlo sampling to approximate choice
probabilities. To avoid these costs, we study decision models. Decision
models replace a language model's text-generation head with a pointer
head that scores candidate options directly (Deußer et al. 2026, Geuter
et al. 2026, Palmer 2026). Unlike LLMs that produce a probability
distributions over all tokens in the vocabulary, decision models produce
a probability distributions over only the valid options in a task. These
probabilities eliminate Monte Carlo sampling cost and yield exact
expected OLS treatment effects. Despite this architectural advantage,
off-the-shelf decision models still incur high treatment-effect error
(Figure~\ref{fig-intro-off-the-shelf}). Because they are trained on
similar single-label objective benchmarks, they treat disagreement among
respondents with the same persona as classification error (Gao et al.
2026).

In this research, we train decision models by updating their weights on
historical human choice microdata. We call this procedure choice
training. Specifically, we minimize respondent-level cross-entropy (CE),
which is equivalent to conditional maximum likelihood estimation. Unlike
post-inference calibration methods that keep model weights fixed and
require human data from the target study (Leng et al. 2026, Wang et al.
2026), choice training occurs before inference and uses no target-study
human data. We therefore ask whether choice training on a small set of
historical datasets can reduce error on held-out test datasets that
never enter training.

We evaluate choice training across three open-weight decision-model
scales (0.8B, 4B, and 9B parameters; Palmer 2026). We train on two
historical choice datasets and select the best checkpoint on a separate
validation dataset. We use four held-out test datasets: a
nutrition-labeling trial (Grummon et al. 2025), a vaccine conjoint
(Kreps et al. 2020), a consumer-rights conjoint (Kantorowicz-Reznichenko
et al. 2026), and a rating conjoint on AI and offshoring (Magistro et
al. 2026). Compared to off-the-shelf decision models, we find that
choice training on two studies lowers both \(\mathrm{MSE}_\beta\) (by
22\% to 89\%) and \(CE\) (by 10\% to 51\%) on every held-out test
dataset and model scale. These gains extend to unseen studies with
attributes and response formats absent from training. We also find that
a small choice-trained model can outperform a much larger off-the-shelf
model on three of the four test datasets.

Our research makes the following contributions to synthetic respondents
in marketing and behavioral research. First, we show that choice
training lowers treatment-effect error on unseen studies without any
target-study human data, complementing existing post-inference
calibration methods (Hullman et al. 2026). Second, we show that training
decision models on just two studies with binary choices transfers to new
domains (vaccines, nutrition labels, consumer rights, AI, and
offshoring) and new response formats (three-option choices and
five-point ratings). Unlike prior research, where training on human
choices failed to generalize to new domains (Brand et al. 2023) or even
within the same study (Wang et al. 2025, Wang et al. 2026), our training
regime transfers to unseen studies. Third, we show that training models
to minimize cross-entropy loss does not guarantee improvements in
treatment effect recovery. Models trained on only one of our training
studies did lower the CE loss on every test dataset, yet sometimes led
to worse treatment-effect error. On the other hand, training on both
studies mitigated these failures. This finding motivates novel training
objectives that target treatment effects directly.

\section{Literature review}\label{sec-lit}

Early work on synthetic respondents prompted off-the-shelf generative
LLMs with demographic profiles to simulate human responses in political,
economic, and marketing surveys (Argyle et al. 2023, Horton et al. 2023,
Santurkar et al. 2023, Goli and Singh 2024). Although these models can
capture average preference directions, systematic discrepancies from
human populations severely bias downstream econometric estimates
(Santurkar et al. 2023, Wadi and Fredette 2025, Leng et al. 2026, Wang
et al. 2026). Common distortions include mode collapse (i.e., producing
nearly the same answer every time a question is asked) and prompt
sensitivity. Recent marketing research addresses these biases while
keeping model weights fixed, relying on human data from the target study
or target individuals. These methods include post-inference econometric
debiasing (Wang et al. 2026), optimal post-inference reweighting across
prompts and checkpoints (Leng et al. 2026), and individual digital twins
conditioned on hundreds of responses per person (Toubia et al. 2025).
Post-inference calibration uses a sample of human responses from the
same study to correct model outputs. Therefore, the final estimates
remain statistically valid even when the model itself is biased.
However, how much these methods gain over human data alone depends on
how closely the underlying model approximates the target human
population (Hullman et al. 2026).

A growing literature instead updates LLM weights directly on human
microdata, either from the target study itself, from a different study
in the same product category, or from many unrelated studies. Training
LLMs on a small sample (e.g., 50 tasks) from the same study cannot
remove the bias in regression parameters (Wang et al. 2025, Ye et al.
2025, Wang et al. 2026) and still requires human data for the same
study. Moreover, research has shown that WTP estimates from a separate
study can transfer to a new study within the same product category, but
fail to generalize to new product categories (Brand et al. 2023).
Outside marketing, large-scale behavioral training improves out-of-study
prediction across cognitive psychology (10 million choices across 160
experiments; {Binz et al.} 2025), social-science experiments (2.9
million responses across 210 studies; Kolluri et al. 2025), and
public-opinion surveys (Suh et al. 2025, Ku et al. 2026, Huang et al.
2026). Yet these cross-study efforts rely on large multi-hundred-study
corpora and autoregressive text generation. Moreover, they evaluate
predictive log-likelihood, marginal distributions, or cell-level
distributional distance rather than econometric treatment-effect
recovery.

A third stream of research tackles the sampling variance of repeated
text generation by using choice probabilities in a single forward pass.
Early studies extract next-token probabilities over option symbols
(e.g., A, B, and C) from LLMs, yet those probabilities remain sensitive
to vocabulary and token-frequency priors ({Kadavath et al.} 2022, Wadi
et al. 2026). Dedicated decision models address this sensitivity by
scoring candidate options outside the vocabulary head (Palmer 2026, Wadi
and Ma 2026). However, because this architecture is new, it remains open
whether pre-inference choice training can adapt decision models for
unseen-study econometric inference without using target-study human
data. In the following section, we formalize our training and evaluation
methodology for these models.

\section{Methodology}\label{sec-methodology}

\begin{figure}

\centering{

\includegraphics[width=0.9\linewidth,height=\textheight,keepaspectratio]{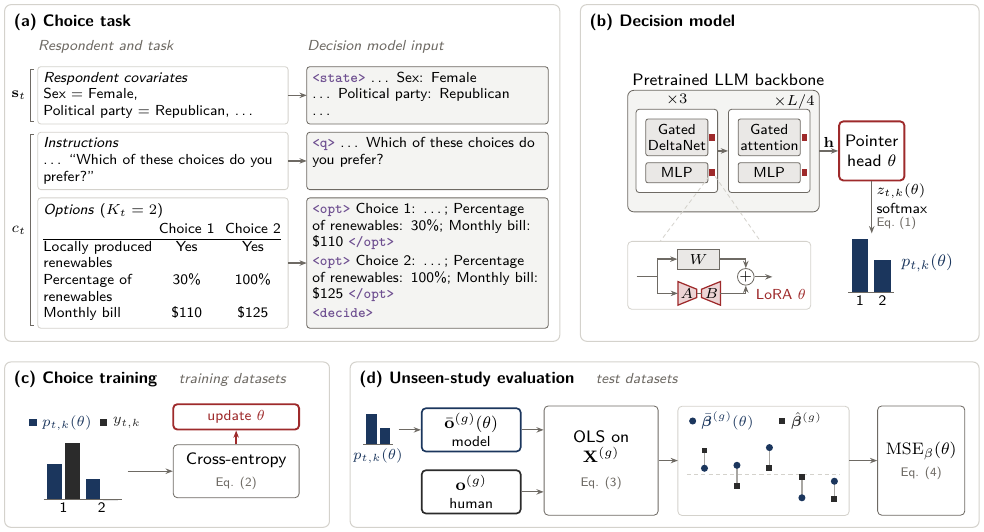}

}

\caption{\label{fig-pipeline}Choice training and unseen-study evaluation
of decision models}

\end{figure}%

\textbf{Decision model parameterization.} Consider a choice study
composed of discrete-choice tasks indexed by \(t\)
(Figure~\ref{fig-pipeline}, panel a). Each task pairs observed
respondent covariates \(\mathbf{s}_t\) (e.g., sex, age, education,
income, political affiliation, and race), which vary by dataset, with a
task context \(c_t\) specifying the instructions and \(K_t \ge 2\)
discrete options. Let \(y_{t,k} \in \{0, 1\}\) indicate whether the
human respondent selects option \(k \in \{1, \dots, K_t\}\) on task
\(t\), with \(\sum_{k=1}^{K_t} y_{t,k} = 1\).

Unlike LLMs that generate tokens autoregressively, a decision model
replaces the vocabulary head of a pretrained language model with a
pointer head that scores all \(K_t\) options in one forward pass
(Figure~\ref{fig-pipeline}, panel b). For each option \(k\), the model
computes a real-valued score \(z_{t,k}(\theta)\) from
\((\mathbf{s}_t, c_t)\). \(\theta\) consists of trainable low-rank
adaptation (LoRA) weights (Hu et al. 2021) on the backbone and the
pointer-head parameters. Applying a softmax transformation with unit
temperature produces the conditional choice probabilities:
\begin{equation}\protect\phantomsection\label{eq-model-pmf}{
p_{t,k}(\theta) = P\big(y_{t,k} = 1 \mid \mathbf{s}_t, c_t; \theta\big) = \frac{\exp\!\big(z_{t,k}(\theta)\big)}{\sum_{m=1}^{K_t} \exp\!\big(z_{t,m}(\theta)\big)}.
}\end{equation} Formally, Equation~\ref{eq-model-pmf} has the
multinomial logit form, with \(z_{t,k}(\theta)\) entering in place of
systematic utility. Because Equation~\ref{eq-model-pmf} normalizes
directly over \(\{1, \dots, K_t\}\) without text decoding, out-of-set
outputs have zero probability by construction.

\textbf{Choice training.} Off-the-shelf decision models are trained on
deterministic tasks with single-label answers. Therefore, the true
conditional distribution of labels places all probability mass on one
option. Minimizing cross-entropy on such data pushes predicted
probabilities \(p_{t,k}(\theta)\) toward zero and one. In contrast, in
human choice studies, respondents with identical observed covariates,
\(\mathbf{s}_t\), still differ in unobserved preferences. Therefore, the
true conditional choice distribution
\(\mathbf{q}_t = (q_{t,1}, \dots, q_{t,K_t})\), with
\(q_{t,k} = P(y_{t,k} = 1 \mid \mathbf{s}_t, c_t)\), spreads probability
mass across options rather than collapsing onto a single mode.

Choice training updates \(\theta\) before inference on historical
respondent-level choice microdata (Figure~\ref{fig-pipeline}, panel c).
Starting from the off-the-shelf model checkpoint, we treat each task as
an independent observation given the respondent's covariates (i.e., a
pooled likelihood), although each respondent may complete several tasks.
We minimize empirical cross-entropy over mini-batches \(S\) of
historical tasks:
\begin{equation}\protect\phantomsection\label{eq-resp-ce}{
\mathcal{L}_{\mathrm{CE}}(\theta; S) = -\frac{1}{|S|} \sum_{t \in S} \sum_{k=1}^{K_t} y_{t,k} \log p_{t,k}(\theta).
}\end{equation} Minimizing Equation~\ref{eq-resp-ce} is equivalent to
pooled conditional maximum likelihood estimation of
Equation~\ref{eq-model-pmf}. This objective is respondent-level because
each term uses an individual respondent's observed choice vector
\(\mathbf{y}_t = (y_{t,1}, \dots, y_{t,K_t})\). Taking expectations over
\(\mathbf{y}_t \sim \mathbf{q}_t\) conditional on
\((\mathbf{s}_t, c_t)\) decomposes the expected task loss as

\begin{equation}\protect\phantomsection\label{eq-ce-kl}{
\begin{aligned}
\mathbb{E}_{\mathbf{y}_t \sim \mathbf{q}_t}\!\left[-\sum_{k=1}^{K_t} y_{t,k} \log p_{t,k}(\theta)\right]
&= -\sum_{k=1}^{K_t} q_{t,k} \log p_{t,k}(\theta) \\
&= H(\mathbf{q}_t) + D_{\mathrm{KL}}\!\big(\mathbf{q}_t \,\|\, \mathbf{p}_t(\theta)\big),
\end{aligned}
}\end{equation}

where
\(\mathbf{p}_t(\theta) = (p_{t,1}(\theta), \dots, p_{t,K_t}(\theta))\)
is the predicted choice distribution from Equation~\ref{eq-model-pmf},
\(H(\mathbf{q}_t) = -\sum_{k=1}^{K_t} q_{t,k} \log q_{t,k}\) is the
Shannon entropy of human choices, and
\(D_{\mathrm{KL}}(\mathbf{q}_t \,\|\, \mathbf{p}_t(\theta)) = \sum_{k=1}^{K_t} q_{t,k} \log\big(q_{t,k} / p_{t,k}(\theta)\big)\)
is the Kullback-Leibler (KL) divergence. Because \(H(\mathbf{q}_t)\)
does not depend on \(\theta\) and
\(D_{\mathrm{KL}}(\mathbf{q}_t \,\|\, \mathbf{p}_t(\theta)) \ge 0\) with
equality if and only if \(\mathbf{p}_t(\theta) = \mathbf{q}_t\),
cross-entropy is a strictly proper scoring rule (i.e., its expected
value is minimized only by the true choice distribution).

Equation~\ref{eq-ce-kl} resolves an apparent tension in our objective
function. Choice training minimizes the same cross-entropy loss on the
same one-hot label format as training regimes that have a single correct
answer. What separates the two is the distribution of \(\mathbf{q}_t\).
On objective tasks, \(\mathbf{q}_t\) places all probability mass on the
correct option (\(H(\mathbf{q}_t) = 0\)), so minimizing the KL term
pushes \(\mathbf{p}_t(\theta)\) toward zero and one. On human choice
data, if respondents with identical \((\mathbf{s}_t, c_t)\) choose
differently, \(\mathbf{q}_t\) becomes interior
(\(H(\mathbf{q}_t) > 0\)). Thus, the same loss function drives
\(\mathbf{p}_t(\theta)\) toward interior probabilities rather than the
most common option. Each one-hot choice is a single draw from
\(\mathbf{q}_t\), hence training requires no aggregate choice shares.
This property is essential in randomized designs with sparse respondent
covariates and task attributes, where each unique combination
\((\mathbf{s}_t, c_t)\) is typically observed only once and, thus,
empirical choice shares cannot be estimated cell by cell. In such cases,
the model recovers \(\mathbf{q}_t\) by sharing parameters across
respondents and tasks with similar covariates and attributes. Because
this argument holds in expectation, a sufficiently flexible model could
still memorize individual choices. Therefore, choice training should
still select the best checkpoint on held-out validation data.
Importantly, no target-study data enters Equation~\ref{eq-resp-ce}, so
downstream econometric evaluation measures genuine generalizability to
unseen studies.

\textbf{Econometric evaluation.} On a held-out evaluation dataset, we
hold \(\theta\) fixed and evaluate treatment-effect recovery using
ordinary least squares (OLS) regressions defined by the dataset's
experimental design (Figure~\ref{fig-pipeline}, panel d). Applying the
same linear specification symmetrically to both human responses and
model-implied expected outcomes places the human benchmark and synthetic
choice data on a common estimator. Consider each regression
\(g \in \{1, \dots, G\}\) in the dataset, with full-column-rank design
matrix \(\mathbf{X}^{(g)}\) and human outcome vector
\(\mathbf{o}^{(g)}\). The matrix \(\mathbf{X}^{(g)}\) contains an
intercept and \(d_g\) treatment or attribute indicators from the
experiment design. Across our evaluation designs, each entry of the
regression outcome \(\mathbf{o}^{(g)}\) is a linear function of the
choice indicators \(y_{t,k}\). It is either a binary indicator
\(y_{t,k}\) for choosing a given profile or target option (zero across
profiles when the outside option is chosen) or the numeric rating
\(\sum_{k=1}^{K_t} k\,y_{t,k}\) on an ordinal scale. By linearity of
expectation, replacing each realized indicator \(y_{t,k}\) with its
predicted probability \(p_{t,k}(\theta)\) from
Equation~\ref{eq-model-pmf} yields the exact expected outcome vector
\(\bar{\mathbf{o}}^{(g)}(\theta)\) that would result from averaging over
synthetic choices drawn from \(\mathbf{p}_t(\theta)\).

Applying the OLS operator
\(\mathbf{A}^{(g)} = (\mathbf{X}^{(g)\top}\mathbf{X}^{(g)})^{-1}\mathbf{X}^{(g)\top}\)
to the human and model-predicted outcome vectors defines the intercepts
and slopes:

\begin{equation}\protect\phantomsection\label{eq-pop-ols}{
\begin{pmatrix}
\hat{\alpha}^{(g)} \\
\hat{\boldsymbol{\beta}}^{(g)}
\end{pmatrix}
= \mathbf{A}^{(g)}\mathbf{o}^{(g)},
\qquad
\begin{pmatrix}
\bar{\alpha}^{(g)}(\theta) \\
\bar{\boldsymbol{\beta}}^{(g)}(\theta)
\end{pmatrix}
= \mathbf{A}^{(g)}\bar{\mathbf{o}}^{(g)}(\theta),
}\end{equation}

where \(\hat{\alpha}^{(g)}\) and \(\bar{\alpha}^{(g)}(\theta)\) are the
intercepts, \(\hat{\boldsymbol{\beta}}^{(g)}\) is the full-sample human
slope benchmark, and \(\bar{\boldsymbol{\beta}}^{(g)}(\theta)\) is the
vector of \(d_g\) synthetic treatment or attribute slopes. Because
\(\mathbf{X}^{(g)}\) is fixed and OLS is linear in the outcome vector,
\(\bar{\boldsymbol{\beta}}^{(g)}(\theta)\) equals the exact expectation
of the OLS slope estimator over synthetic choices drawn from
\(p_{t,k}(\theta)\), with zero Monte Carlo variance.\footnote{Note that
  Equation~\ref{eq-pop-ols} evaluates point estimates rather than OLS
  standard errors, because replacing random choices with expected
  outcomes removes choice-level variance.}

Because all \(G\) regressions within a dataset share identical outcome
units, we measure aggregate slope recovery across all
\(d = \sum_{g=1}^G d_g\) slopes via mean squared error (Wang et al.
2026):

\begin{equation}\protect\phantomsection\label{eq-eval-mse}{
\mathrm{MSE}_{\beta}(\theta) = \frac{1}{d} \sum_{g=1}^G \left\lVert \bar{\boldsymbol{\beta}}^{(g)}(\theta) - \hat{\boldsymbol{\beta}}^{(g)} \right\rVert_2^2.
}\end{equation}

Alongside Equation~\ref{eq-eval-mse}, we evaluate
Equation~\ref{eq-resp-ce} across all tasks in each dataset to report
held-out cross-entropy in nats (Table~\ref{tbl-results-mse}). Whereas
Equation~\ref{eq-eval-mse} measures aggregate slope recovery,
cross-entropy captures how accurately \(\mathbf{p}_t(\theta)\) predicts
conditional choice distributions given \((\mathbf{s}_t, c_t)\)
(Equation~\ref{eq-ce-kl}).

\section{Experimental setup}\label{sec-setup}

\subsection{Datasets}\label{datasets}

For choice training, we use seven datasets from distinct product and
policy domains (Table~\ref{tbl-datasets}). Two datasets provide the
respondent-level training microdata: a paired electricity-plan conjoint
(\texttt{Electricity}; Wu 2025) and a discrete-choice experiment on
drone versus ground delivery across four vendor contexts
(\texttt{Drones}; Berke et al. 2023). Both training datasets vary price
(monthly bill or delivery fee) alongside non-price attributes. A
separate airline conjoint (\texttt{Flights}; Goldfarb et al. 2026)
dataset serves as our validation set, which we use to tune
hyper-parameters and select the best checkpoint (lowest
\(\mathrm{MSE}_\beta\)). Finally, we use four held-out test datasets to
evaluate generalizability to new studies, domains, and choice format.

Two test datasets include prices on products with price ranges absent
from training (\texttt{Consumer\ rights} and
\texttt{AI\ \&\ offshoring}), whereas the other two share no attributes
with the training datasets (\texttt{Food\ labels} and
\texttt{Vaccines}). Response formats in the test sets are also different
from the binary choices used in training (Table~\ref{tbl-datasets}). The
test datasets have a three-option conjoint with an outside option
(\texttt{Vaccines}), a three-option identification task
(\texttt{Food\ labels}), and a five-point ordinal rating
(\texttt{AI\ \&\ offshoring}). Across all seven datasets, each choice
task conditions on the respondent's observed covariates \(\mathbf{s}_t\)
in a standardized text format (Figure~\ref{fig-pipeline}, panel a).

\begin{longtable}[]{@{}
  >{\raggedright\arraybackslash}p{(\linewidth - 12\tabcolsep) * \real{0.2489}}
  >{\raggedright\arraybackslash}p{(\linewidth - 12\tabcolsep) * \real{0.2405}}
  >{\raggedright\arraybackslash}p{(\linewidth - 12\tabcolsep) * \real{0.2511}}
  >{\raggedleft\arraybackslash}p{(\linewidth - 12\tabcolsep) * \real{0.0464}}
  >{\raggedleft\arraybackslash}p{(\linewidth - 12\tabcolsep) * \real{0.0759}}
  >{\raggedleft\arraybackslash}p{(\linewidth - 12\tabcolsep) * \real{0.0865}}
  >{\raggedleft\arraybackslash}p{(\linewidth - 12\tabcolsep) * \real{0.0506}}@{}}
\caption{Summary of training, validation, and held-out test
datasets}\label{tbl-datasets}\tabularnewline
\toprule\noalign{}
\begin{minipage}[b]{\linewidth}\raggedright
Dataset
\end{minipage} & \begin{minipage}[b]{\linewidth}\raggedright
Price attributes
\end{minipage} & \begin{minipage}[b]{\linewidth}\raggedright
Non-price attributes
\end{minipage} & \begin{minipage}[b]{\linewidth}\raggedleft
\(K\)
\end{minipage} & \begin{minipage}[b]{\linewidth}\raggedleft
\(N\)
\end{minipage} & \begin{minipage}[b]{\linewidth}\raggedleft
Tasks
\end{minipage} & \begin{minipage}[b]{\linewidth}\raggedleft
\(d\)
\end{minipage} \\
\midrule\noalign{}
\endfirsthead
\toprule\noalign{}
\begin{minipage}[b]{\linewidth}\raggedright
Dataset
\end{minipage} & \begin{minipage}[b]{\linewidth}\raggedright
Price attributes
\end{minipage} & \begin{minipage}[b]{\linewidth}\raggedright
Non-price attributes
\end{minipage} & \begin{minipage}[b]{\linewidth}\raggedleft
\(K\)
\end{minipage} & \begin{minipage}[b]{\linewidth}\raggedleft
\(N\)
\end{minipage} & \begin{minipage}[b]{\linewidth}\raggedleft
Tasks
\end{minipage} & \begin{minipage}[b]{\linewidth}\raggedleft
\(d\)
\end{minipage} \\
\midrule\noalign{}
\endhead
\bottomrule\noalign{}
\endlastfoot
\emph{Training} & & & & & & \\
Electricity (Wu 2025) & Monthly bill (\$105 to \$125) & Renewable share
(10\% to 100\%), local renewables & 2 & 1,861 & 18,591 & 8 \\
Drones (Berke et al. 2023) & Delivery fee (\$0 to \$10) & Wait time,
drone privacy & 2 & 3,625 & 36,297 & 28 \\
\emph{Validation} & & & & & & \\
Flights (Goldfarb et al. 2026) & Ticket price (\$250 to \$325) &
Sustainable aviation fuel, airline, baggage, legroom, seat & 2 & 1,000 &
5,000 & 12 \\
\emph{Test} & & & & & & \\
Consumer rights (Kantorowicz-Reznichenko et al. 2026) & Laptop (\$500 to
\$2,000) and laundry (\$400 to \$1,400) price & Brand, specs, warranty,
privacy, liability, conflict resolution & 2 & 2,242 & 26,823 & 38 \\
AI \& offshoring (Magistro et al. 2026) & Smartphone price (\$300 to
\$600) & Customer service, factory, and data science jobs & 5 & 5,525 &
21,005 & 32 \\
Food labels (Grummon et al. 2025) & None & Front-of-package warning
labels (6 arms) & 3 & 5,633 & 33,798 & 5 \\
Vaccines (Kreps et al. 2020) & None & Efficacy, duration, side effects,
FDA, origin, endorsements & 3 & 1,971 & 9,855 & 11 \\
\end{longtable}

\emph{Note.} Shorthand labels in the Dataset column are used throughout
the paper. \(K\) is the number of response options per choice task.
\(N\) is the number of respondents. \(d\) reports the total number of
non-reference treatment or attribute slopes across each dataset's
experimental design (Section~\ref{sec-methodology}). Specifically,
\texttt{Consumer\ rights} spans two product conjoints of 19 slopes each
(38 total), and \texttt{AI\ \&\ offshoring} spans four country-by-arm
regressions of 8 slopes each (32 total) on a five-point support scale (1
to 5).

\subsection{Choice-training regimes and
protocol}\label{choice-training-regimes-and-protocol}

At each model scale, we use a \(2 \times 2\) design crossing the
inclusion of \texttt{Electricity} (excluded vs.~included) with the
inclusion of \texttt{Drones} (excluded vs.~included) studies. This forms
four choice-training regimes: \texttt{Off-the-shelf} (neither dataset),
\texttt{Electricity} (18,591 tasks), \texttt{Drones} (36,297 tasks), and
\texttt{Drones\ +\ Electricity} (54,888 tasks). \texttt{Off-the-shelf}
evaluates the unmodified decision model without choice training. This
design tests whether single-dataset training transfers to unseen
datasets and whether pooling both choice datasets improves the model
beyond each individual dataset.

We evaluate the \texttt{KEV-0.8B}, \texttt{KEV-4B}, and \texttt{KEV-9B}
open-weight decision models (Palmer 2026), each loaded in
\texttt{bfloat16} precision with unit temperature. Choice training
updates LoRA adapters (rank \(r = 64\), scaling
\(\alpha_{\mathrm{LoRA}} = 128\), dropout \(0.05\)) across all linear
backbone projections jointly with the trainable pointer head via
Equation~\ref{eq-resp-ce}. Every trained run uses AdamW with a
\(5 \times 10^{-6}\) peak learning rate, a OneCycle schedule (\(10\%\)
warmup and cosine decay), weight decay \(0.01\), and gradient-norm
clipping at \(1.0\), for \(3\) epochs with mini-batch size \(|S| = 40\)
and random seed \(42\). Within each epoch, tasks are shuffled by dataset
(and by vendor context in \texttt{Drones}), partitioned into
single-dataset mini-batches, and interleaved randomly.

The Web Appendix details the off-the-shelf generative LLM protocol
behind Figure~\ref{fig-intro-off-the-shelf}
(\quartoapptblref{apptbl-llm-baseline}), choice-training hyperparameters
(\quartoapptblref{apptbl-hyperparameters}), training curves
(\quartoappfigref{appfig-training-curves}), and attribute-level tables
and diagnostics (\quartoappfigref{appfig-flights-slopes};
\quartoapptblref{apptbl-vaccines-slopes} through
\quartoapptblref{apptbl-magistro-subsets}).

\section{Results}\label{sec-results}

\subsection{Unseen-study transfer across domains and model
scales}\label{unseen-study-transfer-across-domains-and-model-scales}

\begin{figure}

\centering{

\includegraphics[width=1\linewidth,height=\textheight,keepaspectratio]{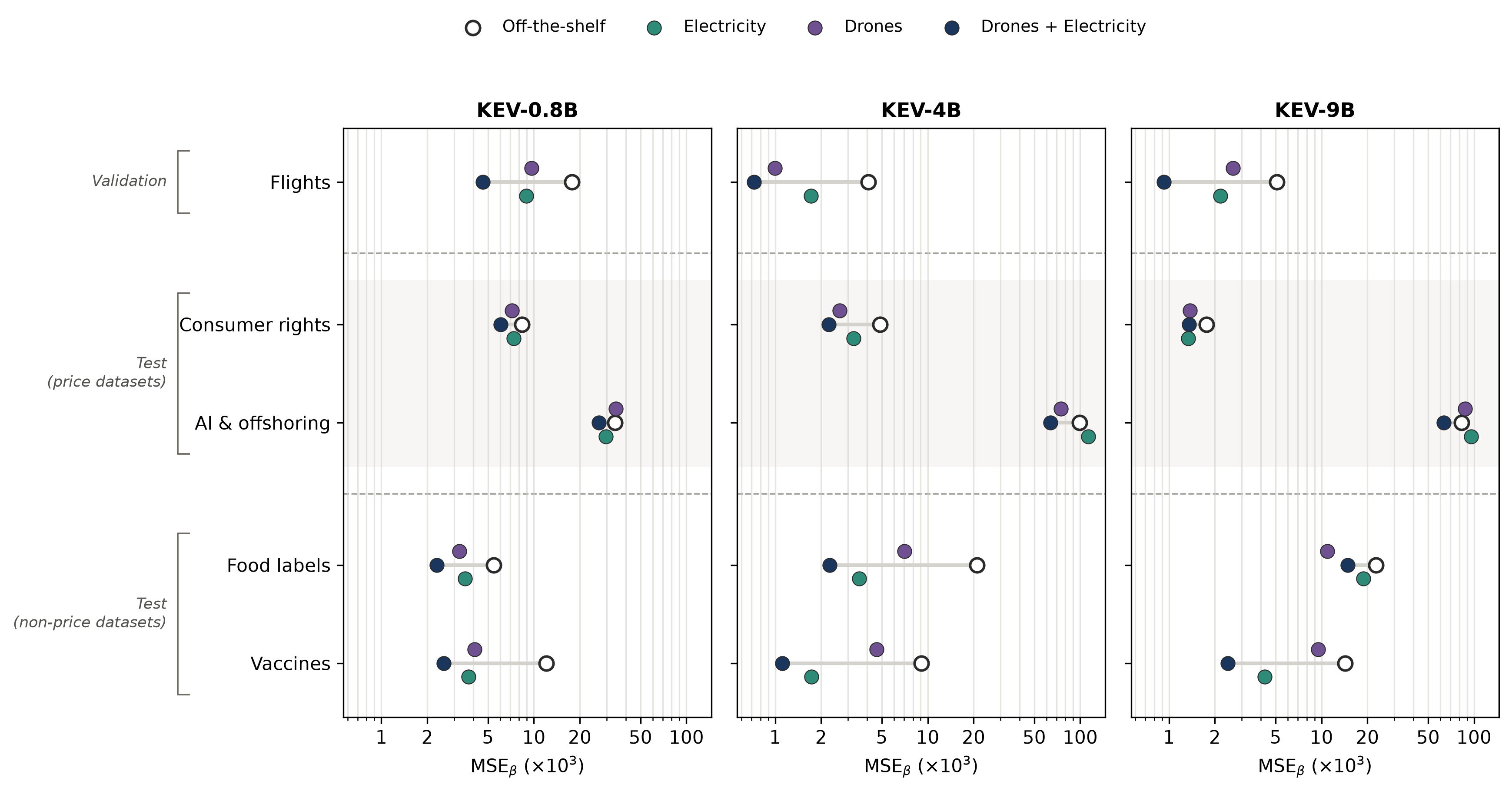}

}

\caption{\label{fig-results-overall}Population OLS treatment-effect
error across datasets and choice-training regimes}

\end{figure}

Relative to the off-the-shelf checkpoints, pooled choice training
(\texttt{Drones\ +\ Electricity}) lowers both \(\mathrm{MSE}_\beta\) and
CE in all 12 held-out test cells, reducing \(\mathrm{MSE}_\beta\) by
22\% to 89\% (Table~\ref{tbl-results-mse}). Notably, these gains extend
both to datasets that share no attributes with the training studies
(\texttt{Food\ labels} and \texttt{Vaccines}) and to response formats
absent from training (Figure~\ref{fig-results-overall}).\footnote{In the
  table, the larger raw \(\mathrm{MSE}_\beta\) values on
  \texttt{AI\ \&\ offshoring} reflect its 1 to 5 ordinal scale rather
  than a binary indicator.} In particular, although the training data
contain only binary choices, pooled models still transfer to
three-option choices and five-point ordinal ratings.

Training on a single dataset, however, transfers to unseen studies less
reliably. Training on \texttt{Electricity} alone and \texttt{Drones}
alone each reduces \(\mathrm{MSE}_\beta\) on three test datasets, yet
raises \(\mathrm{MSE}_\beta\) above off-the-shelf levels in four of six
cells in \texttt{AI\ \&\ offshoring} (Figure~\ref{fig-results-overall}).
\texttt{AI\ \&\ offshoring} is the test dataset that has a five-point
rating scale for its outcome variable. These failures arise even though
single-dataset training lowers CE in all 24 held-out test cells. A
closer approximation of individual choices (i.e., lower CE) therefore
does not guarantee more accurate treatment effects. Pooling both
training datasets eliminates these failures, making unseen-study
transfer more robust than training on either dataset alone.

Choice training also reduces error more reliably than scaling model
parameters. Similar to the pattern seen for generative LLMs
(Figure~\ref{fig-intro-off-the-shelf}), scaling off-the-shelf decision
models fails to improve treatment-effect recovery consistently.
Interestingly, choice-trained \texttt{KEV-0.8B} outperforms
off-the-shelf \texttt{KEV-9B} on three of the four test datasets,
despite having an order of magnitude fewer parameters.

\nexttablesize{\scriptsize}

\begin{longtable}[]{@{}
  >{\raggedright\arraybackslash}p{(\linewidth - 20\tabcolsep) * \real{0.1547}}
  >{\raggedleft\arraybackslash}p{(\linewidth - 20\tabcolsep) * \real{0.0868}}
  >{\raggedleft\arraybackslash}p{(\linewidth - 20\tabcolsep) * \real{0.0755}}
  >{\raggedleft\arraybackslash}p{(\linewidth - 20\tabcolsep) * \real{0.0868}}
  >{\raggedleft\arraybackslash}p{(\linewidth - 20\tabcolsep) * \real{0.0755}}
  >{\raggedleft\arraybackslash}p{(\linewidth - 20\tabcolsep) * \real{0.0868}}
  >{\raggedleft\arraybackslash}p{(\linewidth - 20\tabcolsep) * \real{0.0755}}
  >{\raggedleft\arraybackslash}p{(\linewidth - 20\tabcolsep) * \real{0.0868}}
  >{\raggedleft\arraybackslash}p{(\linewidth - 20\tabcolsep) * \real{0.0755}}
  >{\raggedleft\arraybackslash}p{(\linewidth - 20\tabcolsep) * \real{0.0868}}
  >{\raggedleft\arraybackslash}p{(\linewidth - 20\tabcolsep) * \real{0.0755}}@{}}

\caption{\label{tbl-results-mse}OLS treatment-effect error and empirical
cross-entropy across datasets and choice-training regimes}

\tabularnewline

\toprule\noalign{}
\multirow{2}{=}{\begin{minipage}[b]{\linewidth}\raggedright
Regime
\end{minipage}} &
\multicolumn{2}{>{\centering\arraybackslash}p{(\linewidth - 20\tabcolsep) * \real{0.1623} + 2\tabcolsep}}{%
\begin{minipage}[b]{\linewidth}\centering
Flights
\end{minipage}} &
\multicolumn{2}{>{\centering\arraybackslash}p{(\linewidth - 20\tabcolsep) * \real{0.1623} + 2\tabcolsep}}{%
\begin{minipage}[b]{\linewidth}\centering
Consumer rights
\end{minipage}} &
\multicolumn{2}{>{\centering\arraybackslash}p{(\linewidth - 20\tabcolsep) * \real{0.1623} + 2\tabcolsep}}{%
\begin{minipage}[b]{\linewidth}\centering
AI \& offshoring
\end{minipage}} &
\multicolumn{2}{>{\centering\arraybackslash}p{(\linewidth - 20\tabcolsep) * \real{0.1623} + 2\tabcolsep}}{%
\begin{minipage}[b]{\linewidth}\centering
Food labels
\end{minipage}} &
\multicolumn{2}{>{\centering\arraybackslash}p{(\linewidth - 20\tabcolsep) * \real{0.1623} + 2\tabcolsep}@{}}{%
\begin{minipage}[b]{\linewidth}\centering
Vaccines
\end{minipage}} \\
& \begin{minipage}[b]{\linewidth}\raggedleft
\(\mathrm{MSE}_\beta\)
\end{minipage} & \begin{minipage}[b]{\linewidth}\raggedleft
CE
\end{minipage} & \begin{minipage}[b]{\linewidth}\raggedleft
\(\mathrm{MSE}_\beta\)
\end{minipage} & \begin{minipage}[b]{\linewidth}\raggedleft
CE
\end{minipage} & \begin{minipage}[b]{\linewidth}\raggedleft
\(\mathrm{MSE}_\beta\)
\end{minipage} & \begin{minipage}[b]{\linewidth}\raggedleft
CE
\end{minipage} & \begin{minipage}[b]{\linewidth}\raggedleft
\(\mathrm{MSE}_\beta\)
\end{minipage} & \begin{minipage}[b]{\linewidth}\raggedleft
CE
\end{minipage} & \begin{minipage}[b]{\linewidth}\raggedleft
\(\mathrm{MSE}_\beta\)
\end{minipage} & \begin{minipage}[b]{\linewidth}\raggedleft
CE
\end{minipage} \\
\midrule\noalign{}
\endhead
\bottomrule\noalign{}
\endlastfoot
\multicolumn{11}{@{}>{\centering\arraybackslash}p{(\linewidth - 20\tabcolsep) * \real{0.9660} + 20\tabcolsep}@{}}{%
\emph{KEV-0.8B}} \\
Off-the-shelf & 17.75 & 1.086 & 8.34 & 1.418 & 33.92 & 2.283 & 5.46 &
1.059 & 12.07 & 1.185 \\
Electricity & 8.91 & 0.693 & 7.35 & 0.821 & 29.60 & 1.689 & 3.53 & 0.915
& 3.73 & 1.082 \\
Drones & 9.68 & 0.709 & 7.17 & 0.806 & 34.41 & 1.630 & 3.25 & 0.953 &
4.08 & 1.105 \\
Drones + Electricity & \textbf{4.62} & \textbf{0.647} & \textbf{6.04} &
\textbf{0.695} & \textbf{26.61} & \textbf{1.558} & \textbf{2.30} &
\textbf{0.879} & \textbf{2.56} & \textbf{1.067} \\
\multicolumn{11}{@{}>{\centering\arraybackslash}p{(\linewidth - 20\tabcolsep) * \real{0.9660} + 20\tabcolsep}@{}}{%
\emph{KEV-4B}} \\
Off-the-shelf & 4.10 & 0.811 & 4.89 & 0.779 & 99.01 & 2.434 & 21.10 &
1.100 & 9.08 & 1.525 \\
Electricity & 1.72 & 0.625 & 3.27 & 0.650 & 113.26 & 1.886 & 3.57 &
0.887 & 1.73 & 1.120 \\
Drones & 1.00 & 0.623 & 2.64 & 0.663 & 74.69 & 1.797 & 7.04 & 0.959 &
4.63 & 1.195 \\
Drones + Electricity & \textbf{0.73} & \textbf{0.588} & \textbf{2.25} &
\textbf{0.629} & \textbf{63.98} & \textbf{1.636} & \textbf{2.28} &
\textbf{0.854} & \textbf{1.12} & \textbf{1.046} \\
\multicolumn{11}{@{}>{\centering\arraybackslash}p{(\linewidth - 20\tabcolsep) * \real{0.9660} + 20\tabcolsep}@{}}{%
\emph{KEV-9B}} \\
Off-the-shelf & 5.12 & 0.833 & 1.76 & 0.804 & 82.89 & 2.437 & 22.73 &
1.238 & 14.30 & 1.422 \\
Electricity & 2.17 & 0.647 & \textbf{1.34} & 0.694 & 95.68 & 1.873 &
18.85 & 0.972 & 4.25 & 1.203 \\
Drones & 2.62 & 0.646 & 1.37 & 0.674 & 87.42 & 2.000 & \textbf{10.90} &
0.963 & 9.54 & 1.165 \\
Drones + Electricity & \textbf{0.93} & \textbf{0.604} & 1.36 &
\textbf{0.665} & \textbf{63.49} & \textbf{1.658} & 14.82 &
\textbf{0.808} & \textbf{2.43} & \textbf{1.100} \\

\end{longtable}

\emph{Note.} \(\mathrm{MSE}_\beta\) reports population OLS
treatment-effect error \(\times 10^3\) (Equation~\ref{eq-eval-mse}). CE
reports mean empirical cross-entropy (negative log-likelihood per choice
task, in nats; Equation~\ref{eq-resp-ce}). Lower is better for both
metrics. Bold marks the lowest value per model, dataset, and metric.

\subsection{Attribute-level transfer across shared and unseen
attributes}\label{attribute-level-transfer-across-shared-and-unseen-attributes}

\begin{figure}

\centering{

\includegraphics[width=1\linewidth,height=\textheight,keepaspectratio]{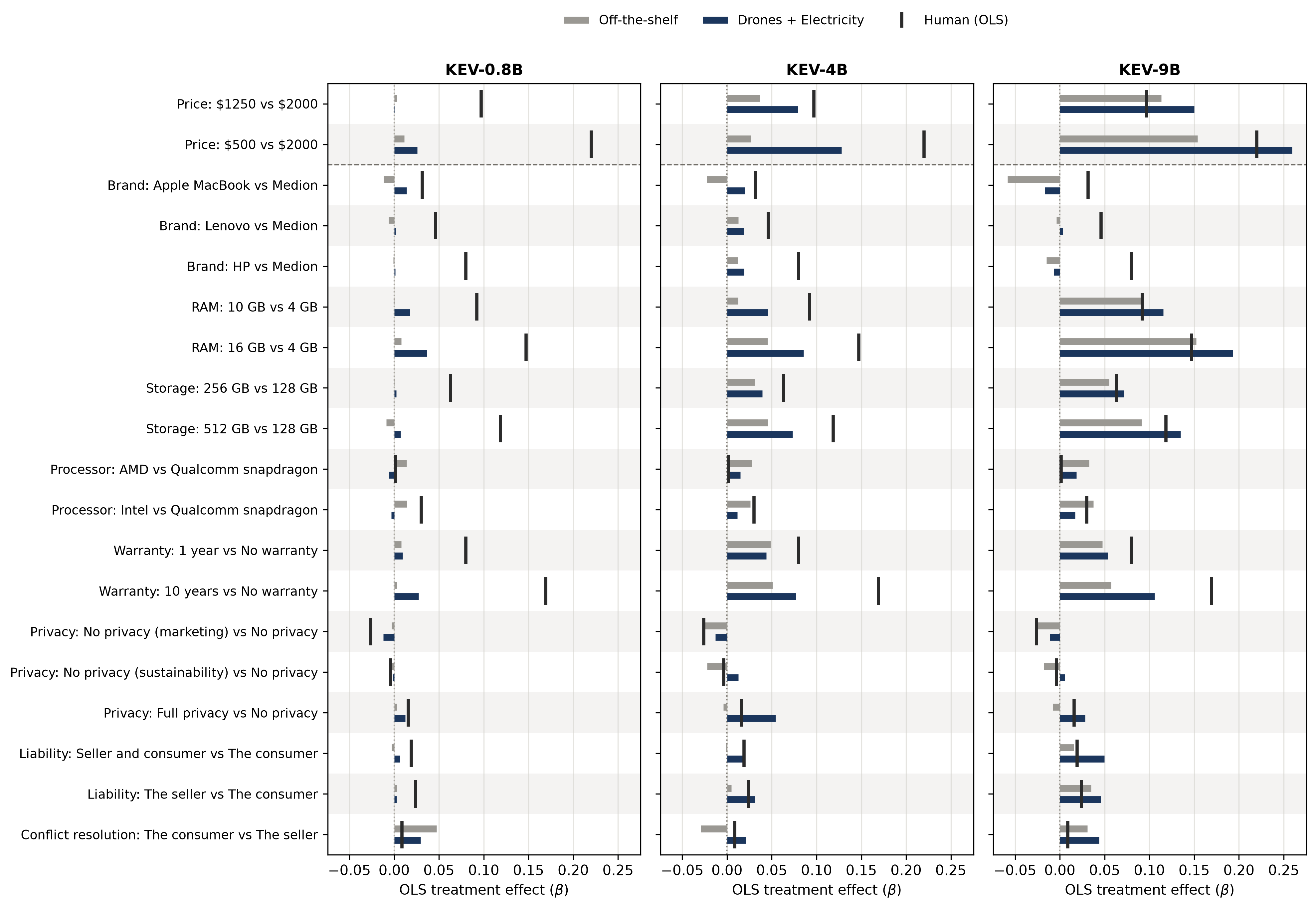}

}

\caption{\label{fig-results-laptops-slopes}Consumer rights (laptops) OLS
treatment effects by attribute for off-the-shelf and pooled
choice-trained models}

\end{figure}

Because both training datasets include price, aggregate gains for the
\texttt{Consumer\ rights} and \texttt{AI\ \&\ offshoring} datasets could
indicate that the models have learned only price-related information.
However, attribute-level estimates rule out this explanation. On
\texttt{Consumer\ rights}, pooled choice training lowers
\(\mathrm{MSE}_\beta\) across the 34 non-price slopes by 23\% to 41\% at
every model scale (Figure~\ref{fig-results-laptops-slopes} and
\quartoapptblref{apptbl-consumer-rights-slopes}). Similarly, it reduces
error across the 24 non-price job slopes on \texttt{AI\ \&\ offshoring},
by 33\% to 34\% at every scale
(\quartoapptblref{apptbl-ai-offshoring-slopes}). Because
\texttt{Food\ labels} and \texttt{Vaccines} share no attributes with
training, their gains likewise cannot stem from price information
learned during training.

\begin{figure}

\centering{

\includegraphics[width=1\linewidth,height=\textheight,keepaspectratio]{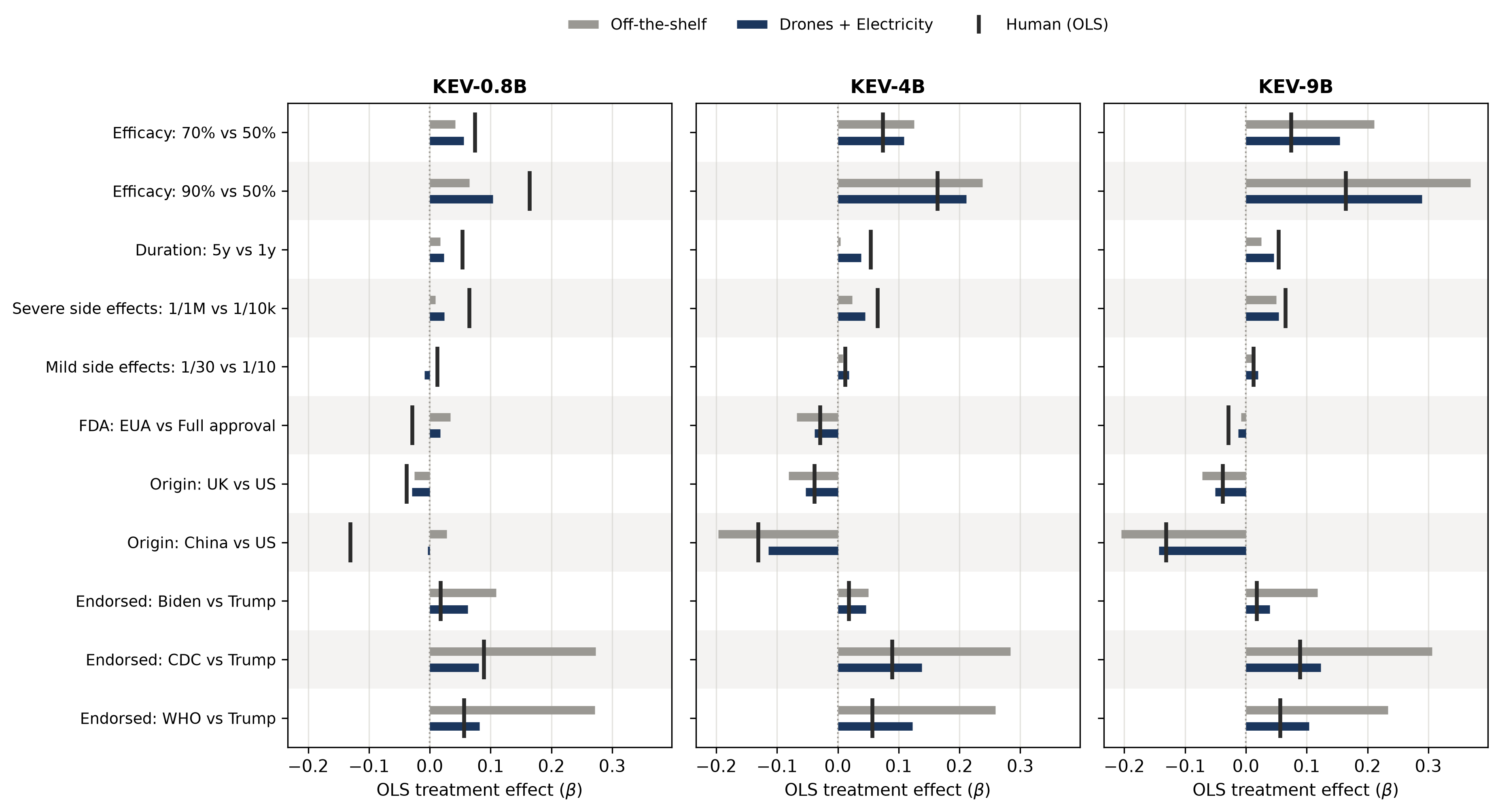}

}

\caption{\label{fig-results-vaccines-slopes}Vaccines OLS treatment
effects by attribute for off-the-shelf and pooled choice-trained models}

\end{figure}

Despite these improvements, pooled choice training narrows but does not
close the gap to human estimates. Some of the largest human effects
remain attenuated or over-estimated at every model scale
(Figure~\ref{fig-results-laptops-slopes} and
Figure~\ref{fig-results-vaccines-slopes}). Full attribute-level
estimates for the remaining datasets appear in the Web Appendix
(\quartoapptblref{apptbl-food-slopes},
\quartoapptblref{apptbl-magistro-subsets},
\quartoappfigref{appfig-flights-slopes}, and
\quartoapptblref{apptbl-flights-slopes}).

\section{Conclusion}\label{sec-conclusion}

Off-the-shelf foundation models, whether generative LLMs or decision
models, struggle to recover human treatment effects in unseen choice
studies. Neither larger parameter counts nor higher reasoning effort
reliably mitigates regression estimates. In this research, we show that
pre-inference choice training of compact open-weight decision models
provides consistent improvements. Training on only two historical choice
studies lowers both OLS treatment-effect error and cross-entropy across
every held-out test dataset and model scale. These gains extend to
unseen attributes and new response formats without any target-study
human data calibration.

Our results show how choice training fits alongside existing approaches
for using AI models as synthetic respondents. Unlike post-inference
calibration methods that require human data from the target study (Leng
et al. 2026, Wang et al. 2026), choice training relies strictly on
historical datasets and can serve as a stronger starting point before
any statistical calibration (Hullman et al. 2026). It also generalizes
beyond the domain of the training datasets and transfers across new
domains and response formats. Finally, whereas large-scale behavioral
models depend on hundreds of studies and autoregressive text generation
({Binz et al.} 2025, Kolluri et al. 2025), our approach achieves
unseen-study transfer from two datasets using a compact model
architecture.

For marketing researchers, these findings carry the following practical
implications. Before any human data is collected, choice-trained
decision models offer a stronger starting point that can support
exploratory pre-testing and experimental design while potentially
reducing the residual bias left for post-inference calibration. In
addition, pooling various training datasets yields more dependable
transfer than single-dataset training. Also, choice training reduces
error more consistently than scaling model parameters. Because decision
models also return exact choice probabilities without Monte Carlo
sampling cost, these gains require far less compute than generative
LLMs.

Our results also expose a limitation of training on cross-entropy alone.
Single-dataset training lowers cross-entropy on every held-out test
cell, yet, for some test datasets, it raises treatment-effect error.
Lower cross-entropy therefore does not guarantee more accurate
experimental slope estimation. One plausible reason is that
cross-entropy weights every choice task equally without directly
penalizing bias in treatment-versus-control contrasts. Because decision
models yield exact, differentiable expected OLS slopes
(Equation~\ref{eq-pop-ols}), future research can train jointly on
respondent-level cross-entropy and a direct treatment-effect loss.

Moreover, because pooled training combines two datasets and more choice
tasks than either single-dataset run, matched-volume ablations are
needed to separate dataset diversity from data volume. Similarly,
choice-training generative LLMs and comparing with decision models would
isolate architectural advantages of decision models from training-data
gains. On the estimation side, choice-trained models still attenuate or
overshoot some of the largest human effects at every scale. Extending
choice training to ordinal scales, non-US samples, and richer respondent
covariates therefore offers a natural next step toward generating more
general synthetic respondents.

\section*{Declarations}\label{declarations}
\addcontentsline{toc}{section}{Declarations}

\subsection*{Funding and competing
interests}\label{funding-and-competing-interests}
\addcontentsline{toc}{subsection}{Funding and competing interests}

All authors certify that they have no affiliations with or involvement
in any organization or entity with any financial interest or
non-financial interest in the subject matter or materials discussed in
this manuscript. The authors have no funding to report.

\section*{References}\label{references}
\addcontentsline{toc}{section}{References}

\protect\phantomsection\label{refs}

\newpage{}

\section*{Web Appendix}\label{web-appendix}
\addcontentsline{toc}{section}{Web Appendix}

\subsection*{Token serialization, query-key readout, and covariate
mappings}\label{token-serialization-query-key-readout-and-covariate-mappings}
\addcontentsline{toc}{subsection}{Token serialization, query-key
readout, and covariate mappings}

\textbf{Serialization.} For each respondent, the covariates
\(\mathbf{s}_t\) are rendered as a state block of ``Label: Value'' lines
(\quartoapptblref{apptbl-covariates}), opened by a state marker. Each
task context \(c_t\) then appears as a branch that begins with a
question marker and the task instructions, places each of the \(K_t\)
options between an option-start and an option-end marker, and ends with
a decision marker (\quartoappfigref{appfig-input-vaccines}). The five
markers reuse reserved tokens from the backbone vocabulary, so the
serialization introduces no new embeddings. Because every task branch
attends to the same shared state block, several tasks for one respondent
can be scored against a single encoded state.

\textbf{Query-key readout.} Let \(\mathbf{h}_t^{\mathrm{dec}}\) be the
final-layer hidden state at the decision marker of task \(t\), and
\(\mathbf{h}_{t,k}\) the hidden state at the option-end marker of option
\(k\). The pointer head projects both into a \(d_p\)-dimensional space,

\begin{equation}\protect\phantomsection\label{eq-pointer-readout}{
\mathbf{a}_t = \mathbf{W}_Q \mathbf{h}_t^{\mathrm{dec}} + \mathbf{c}_Q, \qquad
\mathbf{b}_{t,k} = \mathbf{W}_K \mathbf{h}_{t,k} + \mathbf{c}_K, \qquad
z_{t,k}(\theta) = \frac{\mathbf{b}_{t,k}^{\top} \mathbf{a}_t}{\sqrt{d_p}},
}\end{equation}

with \(d_p = 256\), and Equation~\ref{eq-model-pmf} turns the scores
into choice probabilities. Because the same projections score each
option-end marker separately and Equation~\ref{eq-model-pmf} normalizes
over the listed options, the head handles any number of options \(K_t\).

\nexttablesize{\footnotesize}
{\def\LTcaptype{apptbl}
\begin{longtable}[]{@{}
  >{\raggedright\arraybackslash}p{(\linewidth - 14\tabcolsep) * \real{0.1925}}
  >{\centering\arraybackslash}p{(\linewidth - 14\tabcolsep) * \real{0.1159}}
  >{\centering\arraybackslash}p{(\linewidth - 14\tabcolsep) * \real{0.0911}}
  >{\centering\arraybackslash}p{(\linewidth - 14\tabcolsep) * \real{0.0890}}
  >{\centering\arraybackslash}p{(\linewidth - 14\tabcolsep) * \real{0.1491}}
  >{\centering\arraybackslash}p{(\linewidth - 14\tabcolsep) * \real{0.1491}}
  >{\centering\arraybackslash}p{(\linewidth - 14\tabcolsep) * \real{0.1097}}
  >{\centering\arraybackslash}p{(\linewidth - 14\tabcolsep) * \real{0.1035}}@{}}
\caption{{\label{apptbl-covariates}}Respondent covariates in the
decision model input by dataset}\tabularnewline
\toprule\noalign{}
\begin{minipage}[b]{\linewidth}\raggedright
Covariate
\end{minipage} & \begin{minipage}[b]{\linewidth}\centering
Electricity
\end{minipage} & \begin{minipage}[b]{\linewidth}\centering
Drones
\end{minipage} & \begin{minipage}[b]{\linewidth}\centering
Flights
\end{minipage} & \begin{minipage}[b]{\linewidth}\centering
Consumer rights
\end{minipage} & \begin{minipage}[b]{\linewidth}\centering
AI \& offshoring
\end{minipage} & \begin{minipage}[b]{\linewidth}\centering
Food labels
\end{minipage} & \begin{minipage}[b]{\linewidth}\centering
Vaccines
\end{minipage} \\
\midrule\noalign{}
\endfirsthead
\toprule\noalign{}
\begin{minipage}[b]{\linewidth}\raggedright
Covariate
\end{minipage} & \begin{minipage}[b]{\linewidth}\centering
Electricity
\end{minipage} & \begin{minipage}[b]{\linewidth}\centering
Drones
\end{minipage} & \begin{minipage}[b]{\linewidth}\centering
Flights
\end{minipage} & \begin{minipage}[b]{\linewidth}\centering
Consumer rights
\end{minipage} & \begin{minipage}[b]{\linewidth}\centering
AI \& offshoring
\end{minipage} & \begin{minipage}[b]{\linewidth}\centering
Food labels
\end{minipage} & \begin{minipage}[b]{\linewidth}\centering
Vaccines
\end{minipage} \\
\midrule\noalign{}
\endhead
\bottomrule\noalign{}
\endlastfoot
Sex & ✓ & ✓ & ✓ & ✓ & ✓ & ✓ & ✓ \\
Age & ✓ & ✓ & ✓ & ✓ & ✓ & ✓ & ✓ \\
Education & ✓ & & ✓ & ✓ & ✓ & ✓ & ✓ \\
Income & ✓ & ✓ & ✓ & ✓ & ✓ & ✓ & ✓ \\
Race & ✓ & ✓ & ✓ & & & ✓ & ✓ \\
Hispanic or Latino & ✓ & & & & & ✓ & \\
Region & ✓ & ✓ & ✓ & & ✓ & & ✓ \\
Place of living & & ✓ & ✓ & & & & \\
Religion & & & ✓ & & & & ✓ \\
Political party & ✓ & & ✓ & & ✓ & & ✓ \\
Political ideology & ✓ & & ✓ & & & & ✓ \\
Political self-label & ✓ & & & & & & \\
Voted in last election & & & ✓ & & & & \\
Marital status & ✓ & & & & & & \\
Main activity & ✓ & & & & & & \\
Sexual orientation & ✓ & & & & & & \\
Military service & ✓ & & & & & & \\
Residence type & ✓ & ✓ & & & & & \\
Owns residence & ✓ & & & & & & \\
Rents residence & ✓ & & & & & & \\
Electric heating & ✓ & & & & & & \\
Household size & & & & & & ✓ & \\
Children in household & & & & & & ✓ & \\
SNAP recipient & & & & & & ✓ & \\
WIC recipient & & & & & & ✓ & \\
Nutrition Facts label use & & & & & & ✓ & \\
Online grocery share & & & & & & ✓ & \\
Nutrition literacy score & & & & & & ✓ & \\
\textbf{Total} & \textbf{18} & \textbf{7} & \textbf{11} & \textbf{4} &
\textbf{6} & \textbf{13} & \textbf{9} \\
\end{longtable}
}

\emph{Note.} Each covariate enters the input as one ``Label: Value''
line (\quartoappfigref{appfig-input-vaccines}); covariates a respondent
did not report are omitted. Country is not included.

\begin{appfig}

\centering{

\includegraphics[width=1\linewidth,height=\textheight,keepaspectratio]{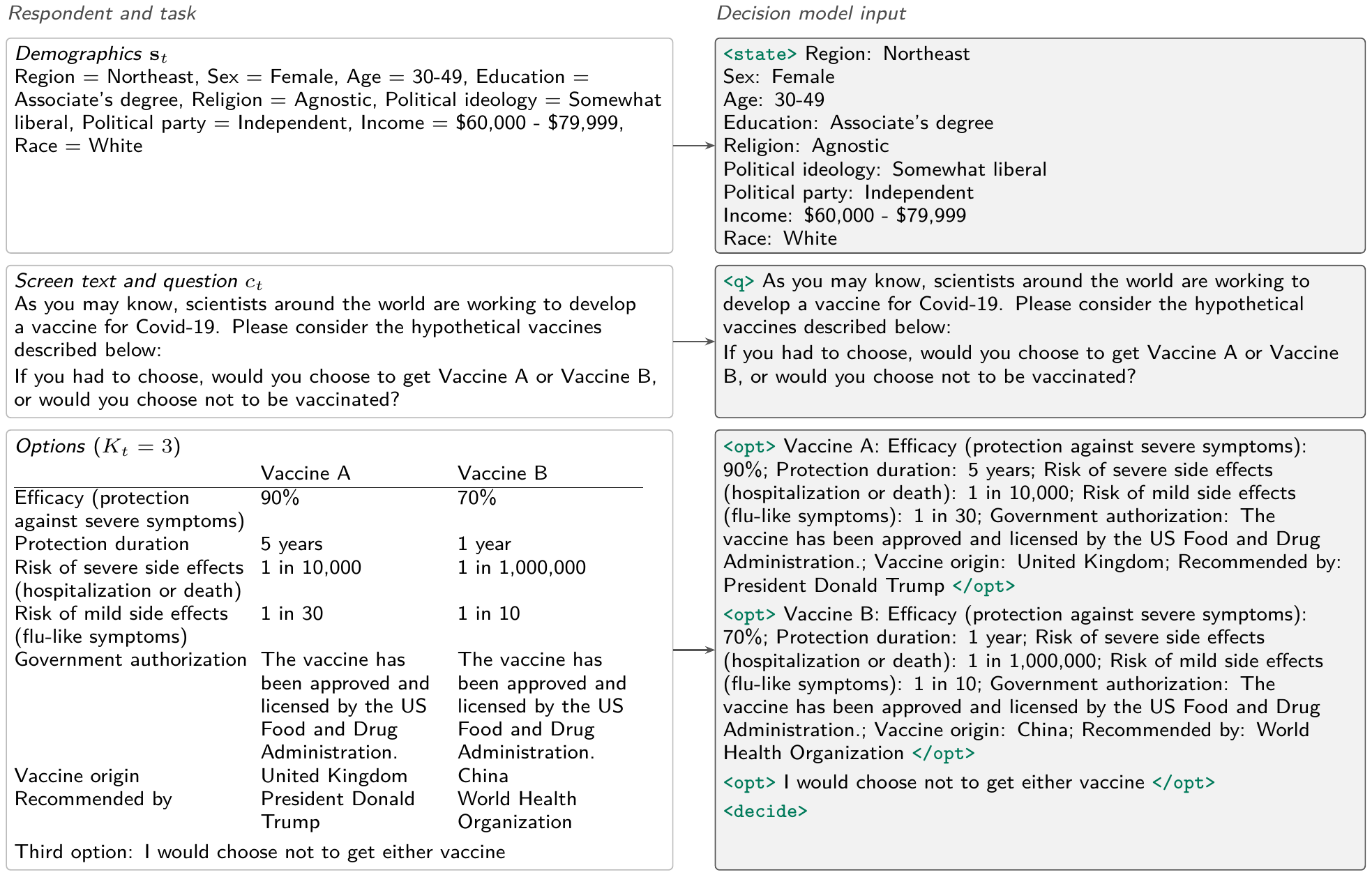}

}

\caption{\label{appfig-input-vaccines}Full decision model input for one
unseen-study task (Vaccines)}

\end{appfig}%

\subsection*{Off-the-shelf generative LLM baseline
protocol}\label{off-the-shelf-generative-llm-baseline-protocol}
\addcontentsline{toc}{subsection}{Off-the-shelf generative LLM baseline
protocol}

We compare off-the-shelf generative LLMs and decision models on a random
subsample of 200 \texttt{Vaccines} respondents (seed 42; 1,000
three-option tasks), matching the AI-only sample size of Wang et al.
(2026). Each model's choices enter the same OLS operator as the main
analysis, and errors are computed against the full-sample human slopes
(\(N = 1,971\); 11 slopes).

\textbf{Prompt.} Generative LLMs receive the prompt of Wang et al.
(2026, sec. 5.2.1). The system prompt reads ``You should act like a
person with the following characteristics deciding whether or not to
receive a COVID-19 vaccination:'', followed by one line per respondent
covariate (e.g., ``Age: 18-29''). The user prompt lists options A and B
by efficacy, protective duration, major and minor side effects,
authorization, origin, and endorsement, adds ``C: You choose neither A
nor B.'', and asks ``Which option would you choose as this person? Your
response should be a single letter A, B, or C.'' Gemini prompts omit the
respondent's region, which the covariate mapping added after those runs.

\textbf{Generation.} We query Gemini 3.5 Flash-Lite (minimal and high
thinking) and Gemini 4 Argon (low and high thinking) through the Gemini
API with default sampling. We serve Qwen3.5-0.8B, 4B, and 9B with vLLM
in \texttt{bfloat16}, toggling reasoning through the chat template and
using Qwen's official recommended sampling settings (reasoning off:
temperature 0.7, top-\(p\) 0.8; reasoning on: temperature 1.0, top-\(p\)
0.95; both top-\(k\) 20 and presence penalty 1.5) without an
output-length limit. Each generative LLM produces one response per task.

\textbf{Parsing.} A response counts as a choice only if, after removing
surrounding whitespace, it is exactly ``A'', ``B'', or ``C''; all other
responses are counted as refusals and excluded from the regression.
Valid responses enter the regression as one-hot choice vectors.

\textbf{Decision models.} Off-the-shelf \texttt{KEV-0.8B},
\texttt{KEV-4B}, and \texttt{KEV-9B} receive the same respondent
covariates and choice task in their native serialization and return a
probability over the three options in one forward pass, which enters the
regression directly. Following the released model cards, these baseline
evaluations apply each checkpoint's shipped calibration temperature
(2.35, 2.41, and 2.19, respectively).

\nexttablesize{\footnotesize}
{\def\LTcaptype{apptbl}
\begin{longtable}[]{@{}
  >{\raggedright\arraybackslash}p{(\linewidth - 8\tabcolsep) * \real{0.3463}}
  >{\raggedright\arraybackslash}p{(\linewidth - 8\tabcolsep) * \real{0.2093}}
  >{\raggedleft\arraybackslash}p{(\linewidth - 8\tabcolsep) * \real{0.1240}}
  >{\raggedleft\arraybackslash}p{(\linewidth - 8\tabcolsep) * \real{0.1163}}
  >{\raggedleft\arraybackslash}p{(\linewidth - 8\tabcolsep) * \real{0.2041}}@{}}
\caption{{\label{apptbl-llm-baseline}}Off-the-shelf generative LLMs and
decision models on the 200-respondent Vaccines subsample}\tabularnewline
\toprule\noalign{}
\begin{minipage}[b]{\linewidth}\raggedright
Model
\end{minipage} & \begin{minipage}[b]{\linewidth}\raggedright
Reasoning effort
\end{minipage} & \begin{minipage}[b]{\linewidth}\raggedleft
Refusals
\end{minipage} & \begin{minipage}[b]{\linewidth}\raggedleft
Opt-out
\end{minipage} & \begin{minipage}[b]{\linewidth}\raggedleft
\(\mathrm{MSE}_\beta\) \(\times 10^3\)
\end{minipage} \\
\midrule\noalign{}
\endfirsthead
\toprule\noalign{}
\begin{minipage}[b]{\linewidth}\raggedright
Model
\end{minipage} & \begin{minipage}[b]{\linewidth}\raggedright
Reasoning effort
\end{minipage} & \begin{minipage}[b]{\linewidth}\raggedleft
Refusals
\end{minipage} & \begin{minipage}[b]{\linewidth}\raggedleft
Opt-out
\end{minipage} & \begin{minipage}[b]{\linewidth}\raggedleft
\(\mathrm{MSE}_\beta\) \(\times 10^3\)
\end{minipage} \\
\midrule\noalign{}
\endhead
\bottomrule\noalign{}
\endlastfoot
Human subsample (200 resp.) & & 0 & 20.3\% & 0.46 \\
Qwen3.5-0.8B & Off & 0 & 59.0\% & 2.75 \\
Qwen3.5-0.8B & On & 19 & 28.6\% & 13.82 \\
Qwen3.5-4B & Off & 0 & 53.6\% & 1.94 \\
Qwen3.5-4B & On & 0 & 4.8\% & 9.71 \\
Qwen3.5-9B & Off & 0 & 81.8\% & 1.49 \\
Qwen3.5-9B & On & 1 & 11.4\% & 6.27 \\
Gemini 3.5 Flash-Lite & Min & 0 & 21.9\% & 10.64 \\
Gemini 3.5 Flash-Lite & High & 0 & 4.8\% & 12.28 \\
Gemini 4 Argon & Low & 0 & 12.2\% & 16.50 \\
Gemini 4 Argon & High & 0 & 9.1\% & 16.57 \\
KEV-0.8B & & 0 & 32.1\% & 5.85 \\
KEV-4B & & 0 & 13.4\% & 2.77 \\
KEV-9B & & 0 & 16.3\% & 4.38 \\
\end{longtable}
}

\emph{Note.} All models condition on respondent demographics over the
same 200-respondent, 1,000-task subsample of \texttt{Vaccines} (Kreps et
al. 2020), evaluated against the full-sample human OLS benchmark
(\(N = 1,971\)). Refusals count tasks where a generative LLM failed to
return a valid choice option. Opt-out reports the share (or expected
probability for decision models) of selecting neither vaccine. Values
back Figure~\ref{fig-intro-off-the-shelf}.

\subsection*{Choice-training
hyperparameters}\label{choice-training-hyperparameters}
\addcontentsline{toc}{subsection}{Choice-training hyperparameters}

\nexttablesize{\footnotesize}
{\def\LTcaptype{apptbl}
\begin{longtable}[]{@{}
  >{\raggedright\arraybackslash}p{(\linewidth - 2\tabcolsep) * \real{0.2979}}
  >{\raggedright\arraybackslash}p{(\linewidth - 2\tabcolsep) * \real{0.7021}}@{}}
\caption{{\label{apptbl-hyperparameters}}Choice-training
hyperparameters shared by all trained runs}\tabularnewline
\toprule\noalign{}
\begin{minipage}[b]{\linewidth}\raggedright
Setting
\end{minipage} & \begin{minipage}[b]{\linewidth}\raggedright
Value
\end{minipage} \\
\midrule\noalign{}
\endfirsthead
\toprule\noalign{}
\begin{minipage}[b]{\linewidth}\raggedright
Setting
\end{minipage} & \begin{minipage}[b]{\linewidth}\raggedright
Value
\end{minipage} \\
\midrule\noalign{}
\endhead
\bottomrule\noalign{}
\endlastfoot
Base decision models & KEV-0.8B (\texttt{bf75a6a}), KEV-4B
(\texttt{6cfce5c}), KEV-9B (\texttt{db029f0}) \\
Weights precision & bfloat16 \\
Pointer-head temperature & 1.0 \\
Pointer-head dimension & 256 (trained in full) \\
LoRA rank \(r\) & 64 \\
LoRA scaling \(\alpha_{\mathrm{LoRA}}\) & 128 \\
LoRA dropout & 0.05 \\
LoRA target modules & All attention, MLP, and Gated DeltaNet
projections \\
Optimizer & AdamW (\(\beta_1 = 0.9\), \(\beta_2 = 0.999\),
\(\epsilon = 10^{-8}\)) \\
Peak learning rate & \(5 \times 10^{-6}\) \\
Learning-rate schedule & OneCycle: 10\% warmup, cosine annealing, start
= peak / 25, end = start / \(10^{4}\) \\
Weight decay & 0.01 \\
Gradient-norm clipping & 1.0 \\
Epochs & 3 \\
Mini-batch size \(\lvert S \rvert\) & 40 tasks from a single dataset \\
Random seed & 42 \\
Checkpoint selection & Lowest Flights \(\mathrm{MSE}_\beta\), evaluated
every 32 steps \\
Hardware & One NVIDIA H100 (80 GB) per run \\
Steps, Electricity & 465 per epoch (1,395 total) \\
Selected step, Electricity & 640 / 416 / 928 \\
Steps, Drones & 910 per epoch (2,730 total) \\
Selected step, Drones & 1,152 / 544 / 256 \\
Steps, Drones + Electricity & 1,375 per epoch (4,125 total) \\
Selected step, Drones + Electricity & 1,600 / 1,472 / 1,600 \\
\end{longtable}
}

\emph{Note.} Settings are identical across the three model scales and
three choice-training regimes. Revision hashes uniquely identify the
base checkpoints of Palmer (2026). Steps count single-dataset
mini-batches. Selected steps report the checkpoint with the lowest
\texttt{Flights} \(\mathrm{MSE}_\beta\) for \texttt{KEV-0.8B},
\texttt{KEV-4B}, and \texttt{KEV-9B}, respectively.

\begin{appfig}

\centering{

\includegraphics[width=1\linewidth,height=\textheight,keepaspectratio]{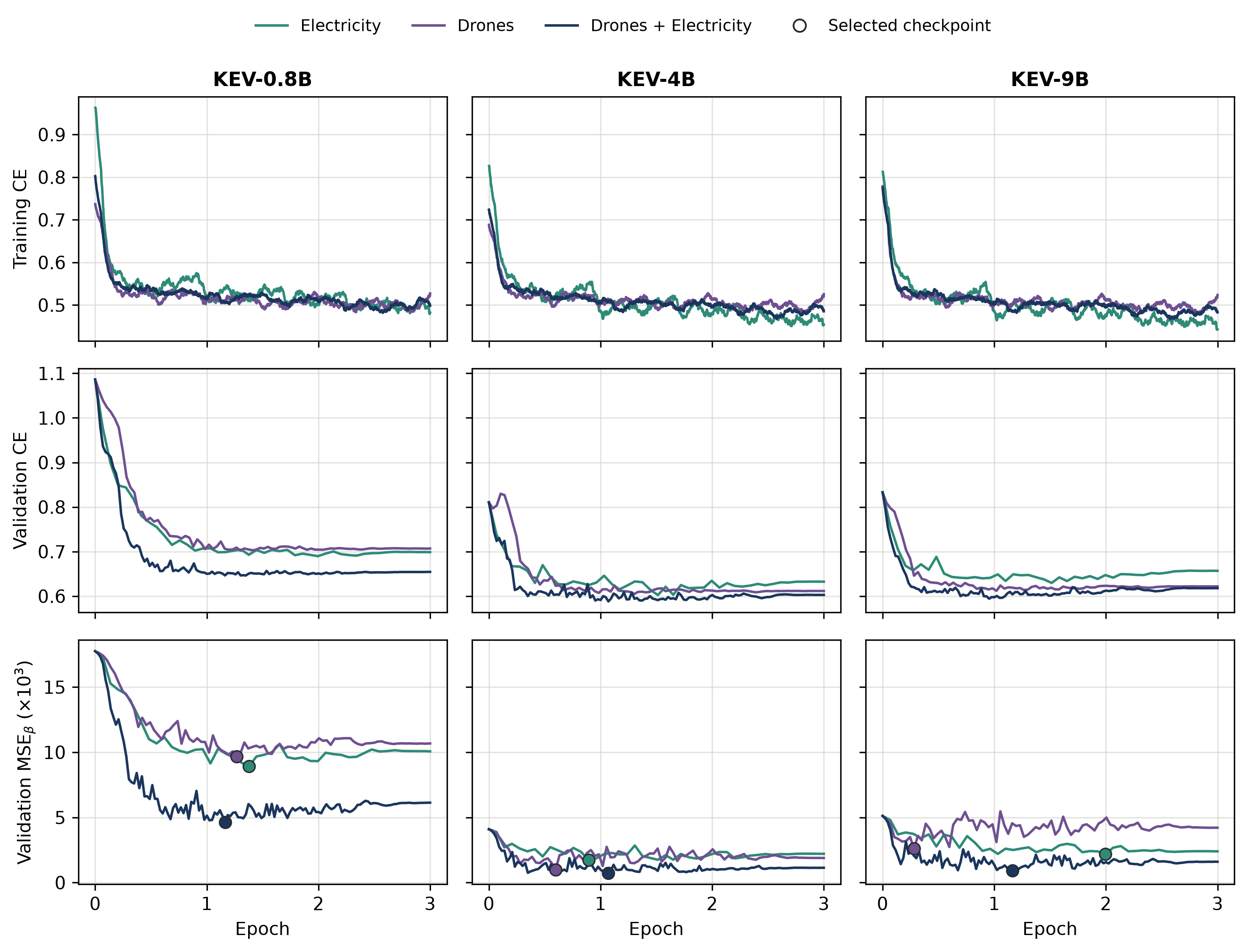}

}

\caption{\label{appfig-training-curves}Training cross-entropy and
validation metrics over choice training}

\floatnote{

\emph{Note.} Training CE is a moving average over 0.1 epoch of the
per-mini-batch loss. Validation metrics are evaluated on
\texttt{Flights} every 32 steps, starting from the off-the-shelf model
at epoch 0. Dots mark the checkpoint with the lowest \texttt{Flights}
\(\mathrm{MSE}_\beta\), which is used for all test evaluations.

}

\end{appfig}%

\subsection*{Attribute-level treatment-effect tables and
diagnostics}\label{attribute-level-treatment-effect-tables-and-diagnostics}
\addcontentsline{toc}{subsection}{Attribute-level treatment-effect
tables and diagnostics}

\begin{appfig}

\centering{

\includegraphics[width=1\linewidth,height=\textheight,keepaspectratio]{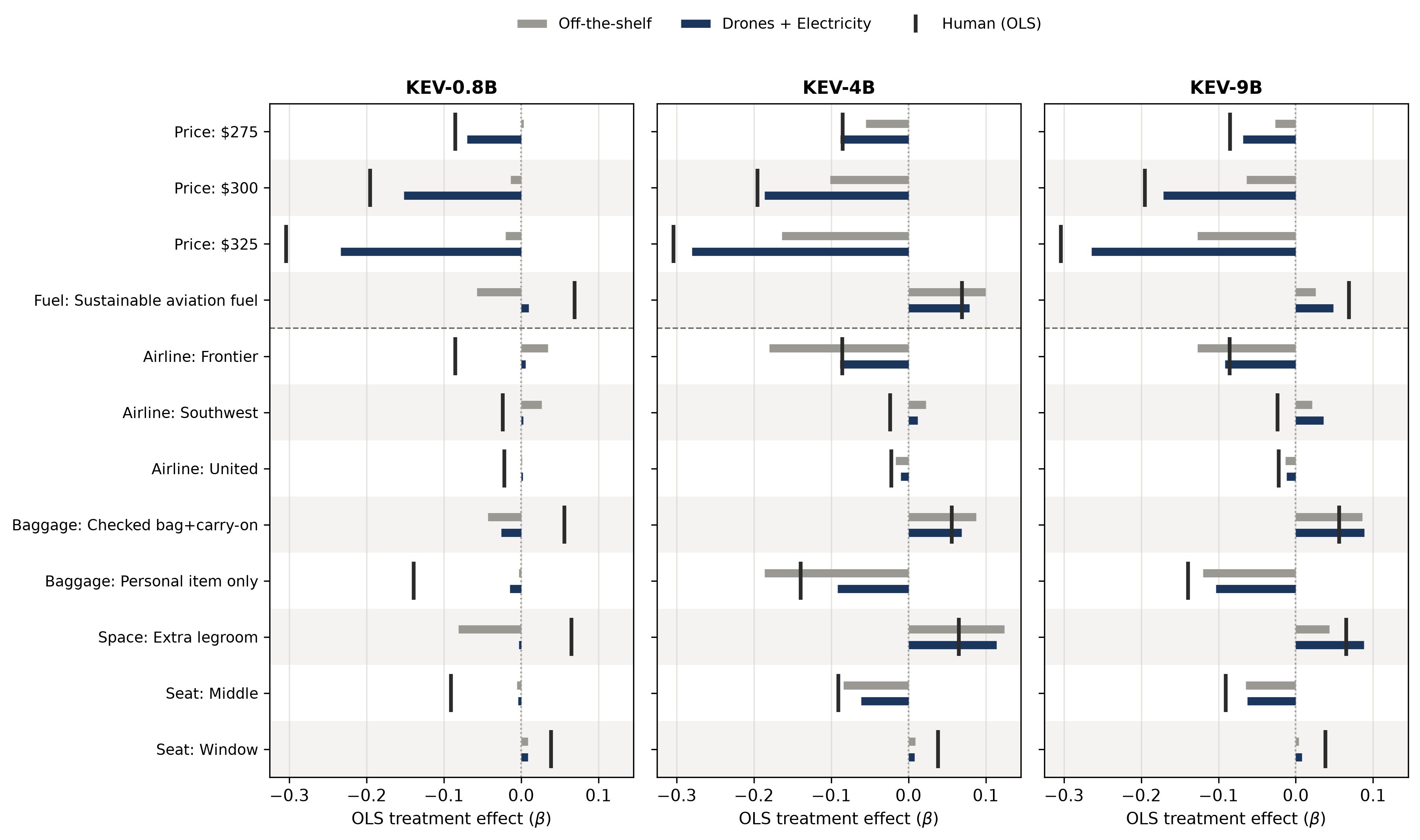}

}

\caption{\label{appfig-flights-slopes}Flights (validation) OLS treatment
effects by attribute for off-the-shelf and pooled choice-trained models}

\end{appfig}%

\nexttablesize{\scriptsize}
{\def\LTcaptype{apptbl}
\begin{longtable}[]{@{}
  >{\raggedright\arraybackslash}p{(\linewidth - 14\tabcolsep) * \real{0.2700}}
  >{\raggedleft\arraybackslash}p{(\linewidth - 14\tabcolsep) * \real{0.0907}}
  >{\raggedleft\arraybackslash}p{(\linewidth - 14\tabcolsep) * \real{0.1076}}
  >{\raggedleft\arraybackslash}p{(\linewidth - 14\tabcolsep) * \real{0.1224}}
  >{\raggedleft\arraybackslash}p{(\linewidth - 14\tabcolsep) * \real{0.0949}}
  >{\raggedleft\arraybackslash}p{(\linewidth - 14\tabcolsep) * \real{0.1097}}
  >{\raggedleft\arraybackslash}p{(\linewidth - 14\tabcolsep) * \real{0.0949}}
  >{\raggedleft\arraybackslash}p{(\linewidth - 14\tabcolsep) * \real{0.1097}}@{}}
\caption{{\label{apptbl-vaccines-slopes}}Vaccines (held-out test) OLS
treatment effects by attribute}\tabularnewline
\toprule\noalign{}
\begin{minipage}[b]{\linewidth}\raggedright
Vaccine attribute level
\end{minipage} & \begin{minipage}[b]{\linewidth}\raggedleft
Human
\end{minipage} & \begin{minipage}[b]{\linewidth}\raggedleft
0.8B Base
\end{minipage} & \begin{minipage}[b]{\linewidth}\raggedleft
0.8B Pooled
\end{minipage} & \begin{minipage}[b]{\linewidth}\raggedleft
4B Base
\end{minipage} & \begin{minipage}[b]{\linewidth}\raggedleft
4B Pooled
\end{minipage} & \begin{minipage}[b]{\linewidth}\raggedleft
9B Base
\end{minipage} & \begin{minipage}[b]{\linewidth}\raggedleft
9B Pooled
\end{minipage} \\
\midrule\noalign{}
\endfirsthead
\toprule\noalign{}
\begin{minipage}[b]{\linewidth}\raggedright
Vaccine attribute level
\end{minipage} & \begin{minipage}[b]{\linewidth}\raggedleft
Human
\end{minipage} & \begin{minipage}[b]{\linewidth}\raggedleft
0.8B Base
\end{minipage} & \begin{minipage}[b]{\linewidth}\raggedleft
0.8B Pooled
\end{minipage} & \begin{minipage}[b]{\linewidth}\raggedleft
4B Base
\end{minipage} & \begin{minipage}[b]{\linewidth}\raggedleft
4B Pooled
\end{minipage} & \begin{minipage}[b]{\linewidth}\raggedleft
9B Base
\end{minipage} & \begin{minipage}[b]{\linewidth}\raggedleft
9B Pooled
\end{minipage} \\
\midrule\noalign{}
\endhead
\bottomrule\noalign{}
\endlastfoot
Efficacy: 70\% vs 50\% & 0.074 & 0.042 & 0.056 & 0.125 & 0.109 & 0.211 &
0.154 \\
Efficacy: 90\% vs 50\% & 0.164 & 0.065 & 0.104 & 0.238 & 0.211 & 0.369 &
0.289 \\
Duration: 5y vs 1y & 0.054 & 0.017 & 0.023 & 0.005 & 0.038 & 0.025 &
0.046 \\
Severe side effects: 1/1M vs 1/10k & 0.065 & 0.009 & 0.024 & 0.024 &
0.045 & 0.050 & 0.054 \\
Mild side effects: 1/30 vs 1/10 & 0.012 & -0.000 & -0.009 & 0.008 &
0.018 & 0.016 & 0.020 \\
FDA: EUA vs Full approval & -0.029 & 0.034 & 0.017 & -0.068 & -0.039 &
-0.008 & -0.013 \\
Origin: UK vs US & -0.038 & -0.025 & -0.030 & -0.081 & -0.053 & -0.072 &
-0.051 \\
Origin: China vs US & -0.131 & 0.028 & -0.004 & -0.197 & -0.114 & -0.205
& -0.143 \\
Endorsed: Biden vs Trump & 0.018 & 0.109 & 0.063 & 0.050 & 0.046 & 0.118
& 0.040 \\
Endorsed: CDC vs Trump & 0.089 & 0.273 & 0.081 & 0.284 & 0.138 & 0.306 &
0.123 \\
Endorsed: WHO vs Trump & 0.057 & 0.271 & 0.082 & 0.259 & 0.123 & 0.234 &
0.104 \\
\textbf{\(\mathrm{MSE}_\beta \times 10^3\)} & 0.00 & \textbf{12.07} &
\textbf{2.56} & \textbf{9.08} & \textbf{1.12} & \textbf{14.30} &
\textbf{2.43} \\
\end{longtable}
}

\emph{Note.} Human reports full-sample OLS slopes on observed respondent
choices in \texttt{Vaccines} (Kreps et al. 2020). Base denotes the
off-the-shelf decision model; Pooled denotes choice training on
\texttt{Drones\ +\ Electricity}. Values back
Figure~\ref{fig-results-vaccines-slopes}.

\nexttablesize{\scriptsize}
{\def\LTcaptype{apptbl}
\begin{longtable}[]{@{}
  >{\raggedright\arraybackslash}p{(\linewidth - 14\tabcolsep) * \real{0.2851}}
  >{\raggedleft\arraybackslash}p{(\linewidth - 14\tabcolsep) * \real{0.0901}}
  >{\raggedleft\arraybackslash}p{(\linewidth - 14\tabcolsep) * \real{0.1048}}
  >{\raggedleft\arraybackslash}p{(\linewidth - 14\tabcolsep) * \real{0.1174}}
  >{\raggedleft\arraybackslash}p{(\linewidth - 14\tabcolsep) * \real{0.0943}}
  >{\raggedleft\arraybackslash}p{(\linewidth - 14\tabcolsep) * \real{0.1069}}
  >{\raggedleft\arraybackslash}p{(\linewidth - 14\tabcolsep) * \real{0.0943}}
  >{\raggedleft\arraybackslash}p{(\linewidth - 14\tabcolsep) * \real{0.1069}}@{}}
\caption{{\label{apptbl-food-slopes}}Food labels (held-out test) OLS
treatment effects by arm}\tabularnewline
\toprule\noalign{}
\begin{minipage}[b]{\linewidth}\raggedright
Warning-label arm
\end{minipage} & \begin{minipage}[b]{\linewidth}\raggedleft
Human
\end{minipage} & \begin{minipage}[b]{\linewidth}\raggedleft
0.8B Base
\end{minipage} & \begin{minipage}[b]{\linewidth}\raggedleft
0.8B Pooled
\end{minipage} & \begin{minipage}[b]{\linewidth}\raggedleft
4B Base
\end{minipage} & \begin{minipage}[b]{\linewidth}\raggedleft
4B Pooled
\end{minipage} & \begin{minipage}[b]{\linewidth}\raggedleft
9B Base
\end{minipage} & \begin{minipage}[b]{\linewidth}\raggedleft
9B Pooled
\end{minipage} \\
\midrule\noalign{}
\endfirsthead
\toprule\noalign{}
\begin{minipage}[b]{\linewidth}\raggedright
Warning-label arm
\end{minipage} & \begin{minipage}[b]{\linewidth}\raggedleft
Human
\end{minipage} & \begin{minipage}[b]{\linewidth}\raggedleft
0.8B Base
\end{minipage} & \begin{minipage}[b]{\linewidth}\raggedleft
0.8B Pooled
\end{minipage} & \begin{minipage}[b]{\linewidth}\raggedleft
4B Base
\end{minipage} & \begin{minipage}[b]{\linewidth}\raggedleft
4B Pooled
\end{minipage} & \begin{minipage}[b]{\linewidth}\raggedleft
9B Base
\end{minipage} & \begin{minipage}[b]{\linewidth}\raggedleft
9B Pooled
\end{minipage} \\
\midrule\noalign{}
\endhead
\bottomrule\noalign{}
\endlastfoot
NutritionInfo vs Positive & 0.184 & 0.128 & 0.161 & 0.250 & 0.199 &
0.358 & 0.344 \\
HighIn vs Positive & 0.054 & 0.123 & 0.076 & -0.208 & -0.020 & -0.124 &
-0.005 \\
Positive+NutritionInfo vs Positive & 0.179 & 0.157 & 0.167 & 0.305 &
0.183 & 0.370 & 0.331 \\
Positive+HighIn vs Positive & 0.119 & -0.008 & 0.023 & -0.002 & 0.047 &
0.019 & 0.192 \\
Spectrum vs Positive & 0.108 & 0.057 & 0.074 & 0.067 & 0.128 & 0.179 &
0.239 \\
\textbf{\(\mathrm{MSE}_\beta \times 10^3\)} & 0.00 & \textbf{5.46} &
\textbf{2.30} & \textbf{21.10} & \textbf{2.28} & \textbf{22.73} &
\textbf{14.82} \\
\end{longtable}
}

\emph{Note.} Human reports full-sample OLS slopes on observed choices in
\texttt{Food\ labels} (Grummon et al. 2025). Base denotes the
off-the-shelf decision model; Pooled denotes choice training on
\texttt{Drones\ +\ Electricity}.

\nexttablesize{\scriptsize}
{\def\LTcaptype{apptbl}
\begin{longtable}[]{@{}
  >{\raggedright\arraybackslash}p{(\linewidth - 14\tabcolsep) * \real{0.3249}}
  >{\raggedleft\arraybackslash}p{(\linewidth - 14\tabcolsep) * \real{0.0901}}
  >{\raggedleft\arraybackslash}p{(\linewidth - 14\tabcolsep) * \real{0.0964}}
  >{\raggedleft\arraybackslash}p{(\linewidth - 14\tabcolsep) * \real{0.1069}}
  >{\raggedleft\arraybackslash}p{(\linewidth - 14\tabcolsep) * \real{0.0901}}
  >{\raggedleft\arraybackslash}p{(\linewidth - 14\tabcolsep) * \real{0.1006}}
  >{\raggedleft\arraybackslash}p{(\linewidth - 14\tabcolsep) * \real{0.0901}}
  >{\raggedleft\arraybackslash}p{(\linewidth - 14\tabcolsep) * \real{0.1006}}@{}}
\caption{{\label{apptbl-consumer-rights-slopes}}Consumer rights
(held-out test) OLS treatment effects by attribute}\tabularnewline
\toprule\noalign{}
\begin{minipage}[b]{\linewidth}\raggedright
Product and attribute level
\end{minipage} & \begin{minipage}[b]{\linewidth}\raggedleft
Human
\end{minipage} & \begin{minipage}[b]{\linewidth}\raggedleft
0.8B Base
\end{minipage} & \begin{minipage}[b]{\linewidth}\raggedleft
0.8B Pooled
\end{minipage} & \begin{minipage}[b]{\linewidth}\raggedleft
4B Base
\end{minipage} & \begin{minipage}[b]{\linewidth}\raggedleft
4B Pooled
\end{minipage} & \begin{minipage}[b]{\linewidth}\raggedleft
9B Base
\end{minipage} & \begin{minipage}[b]{\linewidth}\raggedleft
9B Pooled
\end{minipage} \\
\midrule\noalign{}
\endfirsthead
\toprule\noalign{}
\begin{minipage}[b]{\linewidth}\raggedright
Product and attribute level
\end{minipage} & \begin{minipage}[b]{\linewidth}\raggedleft
Human
\end{minipage} & \begin{minipage}[b]{\linewidth}\raggedleft
0.8B Base
\end{minipage} & \begin{minipage}[b]{\linewidth}\raggedleft
0.8B Pooled
\end{minipage} & \begin{minipage}[b]{\linewidth}\raggedleft
4B Base
\end{minipage} & \begin{minipage}[b]{\linewidth}\raggedleft
4B Pooled
\end{minipage} & \begin{minipage}[b]{\linewidth}\raggedleft
9B Base
\end{minipage} & \begin{minipage}[b]{\linewidth}\raggedleft
9B Pooled
\end{minipage} \\
\midrule\noalign{}
\endhead
\bottomrule\noalign{}
\endlastfoot
Computers \textbar{} Price: \$1250 vs \$2000 & 0.097 & 0.003 & 0.000 &
0.037 & 0.079 & 0.113 & 0.150 \\
Computers \textbar{} Price: \$500 vs \$2000 & 0.220 & 0.011 & 0.026 &
0.027 & 0.128 & 0.154 & 0.259 \\
Computers \textbar{} Brand: Apple MacBook vs Medion & 0.031 & -0.012 &
0.014 & -0.023 & 0.020 & -0.058 & -0.017 \\
Computers \textbar{} Brand: Lenovo vs Medion & 0.046 & -0.006 & 0.002 &
0.013 & 0.019 & -0.004 & 0.003 \\
Computers \textbar{} Brand: HP vs Medion & 0.080 & -0.001 & 0.001 &
0.012 & 0.019 & -0.015 & -0.007 \\
Computers \textbar{} RAM: 10 GB vs 4 GB & 0.092 & -0.001 & 0.018 & 0.012
& 0.046 & 0.093 & 0.116 \\
Computers \textbar{} RAM: 16 GB vs 4 GB & 0.147 & 0.008 & 0.036 & 0.045
& 0.085 & 0.152 & 0.193 \\
Computers \textbar{} Storage: 256 GB vs 128 GB & 0.063 & 0.000 & 0.002 &
0.031 & 0.040 & 0.055 & 0.072 \\
Computers \textbar{} Storage: 512 GB vs 128 GB & 0.119 & -0.009 & 0.007
& 0.046 & 0.073 & 0.092 & 0.135 \\
Computers \textbar{} Processor: AMD vs Qualcomm snapdragon & 0.002 &
0.014 & -0.006 & 0.028 & 0.015 & 0.032 & 0.018 \\
Computers \textbar{} Processor: Intel vs Qualcomm snapdragon & 0.030 &
0.014 & -0.003 & 0.026 & 0.012 & 0.037 & 0.017 \\
Computers \textbar{} Warranty: 1 year vs No warranty & 0.080 & 0.008 &
0.010 & 0.049 & 0.044 & 0.047 & 0.053 \\
Computers \textbar{} Warranty: 10 years vs No warranty & 0.169 & 0.003 &
0.027 & 0.051 & 0.077 & 0.057 & 0.106 \\
Computers \textbar{} Privacy: No privacy (marketing) vs No privacy &
-0.026 & -0.003 & -0.012 & -0.028 & -0.013 & -0.025 & -0.011 \\
Computers \textbar{} Privacy: No privacy (sustainability) vs No privacy
& -0.004 & -0.003 & -0.002 & -0.022 & 0.013 & -0.018 & 0.006 \\
Computers \textbar{} Privacy: Full privacy vs No privacy & 0.016 & 0.003
& 0.012 & -0.004 & 0.054 & -0.008 & 0.028 \\
Computers \textbar{} Liability: Seller and consumer vs The consumer &
0.019 & -0.003 & 0.006 & -0.001 & 0.020 & 0.016 & 0.050 \\
Computers \textbar{} Liability: The seller vs The consumer & 0.024 &
0.003 & 0.003 & 0.005 & 0.031 & 0.035 & 0.046 \\
Computers \textbar{} Conflict resolution: The consumer vs The seller &
0.009 & 0.047 & 0.029 & -0.029 & 0.021 & 0.031 & 0.044 \\
Laundry \textbar{} Price: \$900 vs \$1400 & 0.113 & 0.012 & 0.045 &
0.035 & 0.072 & 0.097 & 0.132 \\
Laundry \textbar{} Price: \$400 vs \$1400 & 0.215 & 0.011 & 0.092 &
0.056 & 0.127 & 0.179 & 0.269 \\
Laundry \textbar{} Brand: Hoover vs Haier & 0.018 & -0.010 & -0.015 &
0.004 & 0.004 & -0.006 & 0.002 \\
Laundry \textbar{} Brand: General Electric vs Haier & 0.043 & 0.002 &
-0.014 & 0.007 & 0.008 & 0.011 & 0.008 \\
Laundry \textbar{} Brand: Whirlpool vs Haier & 0.061 & 0.000 & -0.009 &
0.011 & 0.009 & 0.011 & 0.010 \\
Laundry \textbar{} Depth: 32 inches vs 28 inches & 0.033 & 0.001 & 0.006
& -0.002 & -0.002 & -0.010 & -0.008 \\
Laundry \textbar{} Depth: 36 inches vs 28 inches & 0.055 & -0.001 &
0.008 & -0.004 & -0.004 & -0.015 & -0.015 \\
Laundry \textbar{} Noise: 75 dB vs 78 dB & 0.005 & -0.013 & 0.006 &
0.004 & 0.022 & 0.019 & 0.035 \\
Laundry \textbar{} Noise: 70 dB vs 78 dB & 0.019 & -0.010 & 0.021 &
0.018 & 0.050 & 0.046 & 0.079 \\
Laundry \textbar{} Load capacity: 12 pounds vs 6 pounds & 0.096 & 0.005
& 0.009 & 0.041 & 0.055 & 0.105 & 0.114 \\
Laundry \textbar{} Load capacity: 20 pounds vs 6 pounds & 0.170 & -0.005
& 0.016 & 0.067 & 0.092 & 0.150 & 0.181 \\
Laundry \textbar{} Warranty: 1 year vs No warranty & 0.103 & -0.006 &
0.009 & 0.043 & 0.040 & 0.065 & 0.071 \\
Laundry \textbar{} Warranty: 10 years vs No warranty & 0.227 & -0.003 &
0.029 & 0.065 & 0.092 & 0.134 & 0.171 \\
Laundry \textbar{} Privacy: No privacy (marketing) vs No privacy &
-0.013 & 0.008 & -0.004 & -0.037 & -0.006 & -0.017 & -0.005 \\
Laundry \textbar{} Privacy: No privacy (sustainability) vs No privacy &
0.010 & 0.003 & 0.006 & -0.022 & 0.019 & -0.018 & 0.005 \\
Laundry \textbar{} Privacy: Full privacy vs No privacy & 0.022 & 0.026 &
-0.003 & -0.011 & 0.049 & -0.008 & 0.026 \\
Laundry \textbar{} Liability: Seller and consumer vs The consumer &
0.024 & 0.012 & 0.014 & -0.019 & 0.004 & 0.006 & 0.048 \\
Laundry \textbar{} Liability: The seller vs The consumer & 0.033 & 0.020
& 0.009 & -0.019 & 0.017 & 0.025 & 0.043 \\
Laundry \textbar{} Conflict resolution: The consumer vs The seller &
0.020 & 0.055 & 0.034 & -0.040 & 0.015 & 0.025 & 0.043 \\
\textbf{Price \(\mathrm{MSE}_\beta \times 10^3\) (4)} & 0.00 &
\textbf{26.02} & \textbf{16.64} & \textbf{18.12} & \textbf{4.54} &
\textbf{1.53} & \textbf{1.91} \\
\textbf{Non-price \(\mathrm{MSE}_\beta \times 10^3\) (34)} & 0.00 &
\textbf{6.26} & \textbf{4.79} & \textbf{3.33} & \textbf{1.98} &
\textbf{1.79} & \textbf{1.29} \\
\textbf{Overall \(\mathrm{MSE}_\beta \times 10^3\) (38)} & 0.00 &
\textbf{8.34} & \textbf{6.04} & \textbf{4.89} & \textbf{2.25} &
\textbf{1.76} & \textbf{1.36} \\
\end{longtable}
}

\emph{Note.} Human reports full-sample OLS slopes across the 19 laptop
(\texttt{Computers}) and 19 laundry-machine (\texttt{Laundry}) attribute
levels in \texttt{Consumer\ rights} (Kantorowicz-Reznichenko et al.
2026). Base denotes the off-the-shelf decision model; Pooled denotes
choice training on \texttt{Drones\ +\ Electricity}. Laptop slopes back
Figure~\ref{fig-results-laptops-slopes}.

\nexttablesize{\scriptsize}
{\def\LTcaptype{apptbl}
\begin{longtable}[]{@{}
  >{\raggedright\arraybackslash}p{(\linewidth - 14\tabcolsep) * \real{0.3095}}
  >{\raggedleft\arraybackslash}p{(\linewidth - 14\tabcolsep) * \real{0.0905}}
  >{\raggedleft\arraybackslash}p{(\linewidth - 14\tabcolsep) * \real{0.0989}}
  >{\raggedleft\arraybackslash}p{(\linewidth - 14\tabcolsep) * \real{0.1095}}
  >{\raggedleft\arraybackslash}p{(\linewidth - 14\tabcolsep) * \real{0.0926}}
  >{\raggedleft\arraybackslash}p{(\linewidth - 14\tabcolsep) * \real{0.1032}}
  >{\raggedleft\arraybackslash}p{(\linewidth - 14\tabcolsep) * \real{0.0926}}
  >{\raggedleft\arraybackslash}p{(\linewidth - 14\tabcolsep) * \real{0.1032}}@{}}
\caption{{\label{apptbl-ai-offshoring-slopes}}AI \& offshoring
(held-out test) OLS treatment effects by attribute}\tabularnewline
\toprule\noalign{}
\begin{minipage}[b]{\linewidth}\raggedright
Country, arm, and attribute level
\end{minipage} & \begin{minipage}[b]{\linewidth}\raggedleft
Human
\end{minipage} & \begin{minipage}[b]{\linewidth}\raggedleft
0.8B Base
\end{minipage} & \begin{minipage}[b]{\linewidth}\raggedleft
0.8B Pooled
\end{minipage} & \begin{minipage}[b]{\linewidth}\raggedleft
4B Base
\end{minipage} & \begin{minipage}[b]{\linewidth}\raggedleft
4B Pooled
\end{minipage} & \begin{minipage}[b]{\linewidth}\raggedleft
9B Base
\end{minipage} & \begin{minipage}[b]{\linewidth}\raggedleft
9B Pooled
\end{minipage} \\
\midrule\noalign{}
\endfirsthead
\toprule\noalign{}
\begin{minipage}[b]{\linewidth}\raggedright
Country, arm, and attribute level
\end{minipage} & \begin{minipage}[b]{\linewidth}\raggedleft
Human
\end{minipage} & \begin{minipage}[b]{\linewidth}\raggedleft
0.8B Base
\end{minipage} & \begin{minipage}[b]{\linewidth}\raggedleft
0.8B Pooled
\end{minipage} & \begin{minipage}[b]{\linewidth}\raggedleft
4B Base
\end{minipage} & \begin{minipage}[b]{\linewidth}\raggedleft
4B Pooled
\end{minipage} & \begin{minipage}[b]{\linewidth}\raggedleft
9B Base
\end{minipage} & \begin{minipage}[b]{\linewidth}\raggedleft
9B Pooled
\end{minipage} \\
\midrule\noalign{}
\endhead
\bottomrule\noalign{}
\endlastfoot
US AI \textbar{} Price: \$450 vs \$600 & 0.313 & -0.002 & 0.031 & 0.105
& 0.171 & 0.236 & 0.331 \\
US AI \textbar{} Price: \$300 vs \$600 & 0.351 & -0.013 & 0.022 & 0.174
& 0.242 & 0.256 & 0.364 \\
US AI \textbar{} Customer service jobs: 75 vs 100 & -0.197 & -0.010 &
-0.054 & -0.296 & -0.251 & -0.121 & -0.154 \\
US AI \textbar{} Customer service jobs: 125 vs 100 & 0.072 & 0.037 &
0.061 & 0.394 & 0.294 & 0.172 & 0.153 \\
US AI \textbar{} Factory jobs: 300 vs 400 & -0.389 & -0.078 & -0.125 &
-0.502 & -0.590 & -0.511 & -0.613 \\
US AI \textbar{} Factory jobs: 500 vs 400 & 0.140 & 0.028 & 0.072 &
0.800 & 0.638 & 0.841 & 0.636 \\
US AI \textbar{} Data science jobs: 150 vs 200 & -0.215 & -0.055 &
-0.087 & -0.408 & -0.376 & -0.258 & -0.347 \\
US AI \textbar{} Data science jobs: 250 vs 200 & 0.075 & 0.109 & 0.160 &
0.329 & 0.248 & 0.020 & -0.001 \\
US Offshoring \textbar{} Price: \$450 vs \$600 & 0.227 & 0.005 & 0.016 &
0.141 & 0.186 & 0.414 & 0.522 \\
US Offshoring \textbar{} Price: \$300 vs \$600 & 0.261 & -0.032 & -0.023
& 0.159 & 0.214 & 0.466 & 0.587 \\
US Offshoring \textbar{} Customer service jobs: 75 vs 100 & -0.185 &
-0.010 & -0.059 & -0.351 & -0.249 & -0.145 & -0.146 \\
US Offshoring \textbar{} Customer service jobs: 125 vs 100 & 0.108 &
0.045 & 0.062 & 0.352 & 0.242 & 0.207 & 0.194 \\
US Offshoring \textbar{} Factory jobs: 300 vs 400 & -0.308 & -0.038 &
-0.091 & -0.442 & -0.495 & -0.509 & -0.552 \\
US Offshoring \textbar{} Factory jobs: 500 vs 400 & 0.081 & 0.067 &
0.115 & 0.787 & 0.652 & 0.800 & 0.576 \\
US Offshoring \textbar{} Data science jobs: 150 vs 200 & -0.128 & -0.015
& -0.069 & -0.297 & -0.273 & -0.262 & -0.307 \\
US Offshoring \textbar{} Data science jobs: 250 vs 200 & 0.087 & 0.142 &
0.199 & 0.422 & 0.333 & 0.069 & 0.071 \\
Canada AI \textbar{} Price: \$450 vs \$600 & 0.209 & -0.005 & 0.022 &
0.118 & 0.163 & 0.214 & 0.303 \\
Canada AI \textbar{} Price: \$300 vs \$600 & 0.263 & -0.020 & 0.007 &
0.163 & 0.219 & 0.225 & 0.322 \\
Canada AI \textbar{} Customer service jobs: 75 vs 100 & -0.188 & -0.015
& -0.060 & -0.297 & -0.240 & -0.116 & -0.166 \\
Canada AI \textbar{} Customer service jobs: 125 vs 100 & 0.036 & 0.038 &
0.067 & 0.339 & 0.281 & 0.149 & 0.139 \\
Canada AI \textbar{} Factory jobs: 300 vs 400 & -0.251 & -0.076 & -0.132
& -0.490 & -0.567 & -0.488 & -0.608 \\
Canada AI \textbar{} Factory jobs: 500 vs 400 & 0.111 & 0.036 & 0.083 &
0.769 & 0.649 & 0.841 & 0.650 \\
Canada AI \textbar{} Data science jobs: 150 vs 200 & -0.069 & -0.049 &
-0.084 & -0.410 & -0.356 & -0.236 & -0.321 \\
Canada AI \textbar{} Data science jobs: 250 vs 200 & 0.077 & 0.112 &
0.165 & 0.292 & 0.230 & -0.005 & -0.028 \\
Canada Offshoring \textbar{} Price: \$450 vs \$600 & 0.195 & 0.006 &
0.011 & 0.139 & 0.171 & 0.394 & 0.493 \\
Canada Offshoring \textbar{} Price: \$300 vs \$600 & 0.274 & -0.029 &
-0.027 & 0.154 & 0.207 & 0.468 & 0.606 \\
Canada Offshoring \textbar{} Customer service jobs: 75 vs 100 & -0.156 &
-0.013 & -0.063 & -0.334 & -0.244 & -0.136 & -0.148 \\
Canada Offshoring \textbar{} Customer service jobs: 125 vs 100 & 0.009 &
0.040 & 0.066 & 0.346 & 0.243 & 0.229 & 0.215 \\
Canada Offshoring \textbar{} Factory jobs: 300 vs 400 & -0.264 & -0.036
& -0.094 & -0.404 & -0.476 & -0.493 & -0.527 \\
Canada Offshoring \textbar{} Factory jobs: 500 vs 400 & 0.050 & 0.069 &
0.122 & 0.806 & 0.682 & 0.836 & 0.622 \\
Canada Offshoring \textbar{} Data science jobs: 150 vs 200 & -0.164 &
-0.012 & -0.068 & -0.304 & -0.269 & -0.233 & -0.260 \\
Canada Offshoring \textbar{} Data science jobs: 250 vs 200 & 0.073 &
0.135 & 0.200 & 0.417 & 0.325 & 0.105 & 0.090 \\
\textbf{Price \(\mathrm{MSE}_\beta \times 10^3\) (8)} & 0.00 &
\textbf{77.60} & \textbf{67.37} & \textbf{16.09} & \textbf{5.66} &
\textbf{21.31} & \textbf{50.56} \\
\textbf{Non-price \(\mathrm{MSE}_\beta \times 10^3\) (24)} & 0.00 &
\textbf{19.36} & \textbf{13.03} & \textbf{126.66} & \textbf{83.42} &
\textbf{103.41} & \textbf{67.80} \\
\textbf{Overall \(\mathrm{MSE}_\beta \times 10^3\) (32)} & 0.00 &
\textbf{33.92} & \textbf{26.61} & \textbf{99.01} & \textbf{63.98} &
\textbf{82.89} & \textbf{63.49} \\
\end{longtable}
}

\emph{Note.} Human reports full-sample OLS slopes on the five-point
ordinal support scale across four country-by-arm regressions (8 slopes
each, 32 total) in \texttt{AI\ \&\ offshoring} (Magistro et al. 2026).
Base denotes the off-the-shelf decision model; Pooled denotes choice
training on \texttt{Drones\ +\ Electricity}.

\nexttablesize{\scriptsize}
{\def\LTcaptype{apptbl}
\begin{longtable}[]{@{}
  >{\raggedright\arraybackslash}p{(\linewidth - 14\tabcolsep) * \real{0.2647}}
  >{\raggedleft\arraybackslash}p{(\linewidth - 14\tabcolsep) * \real{0.0903}}
  >{\raggedleft\arraybackslash}p{(\linewidth - 14\tabcolsep) * \real{0.1071}}
  >{\raggedleft\arraybackslash}p{(\linewidth - 14\tabcolsep) * \real{0.1218}}
  >{\raggedleft\arraybackslash}p{(\linewidth - 14\tabcolsep) * \real{0.0966}}
  >{\raggedleft\arraybackslash}p{(\linewidth - 14\tabcolsep) * \real{0.1113}}
  >{\raggedleft\arraybackslash}p{(\linewidth - 14\tabcolsep) * \real{0.0966}}
  >{\raggedleft\arraybackslash}p{(\linewidth - 14\tabcolsep) * \real{0.1113}}@{}}
\caption{{\label{apptbl-flights-slopes}}Flights (validation) OLS
treatment effects by attribute}\tabularnewline
\toprule\noalign{}
\begin{minipage}[b]{\linewidth}\raggedright
Flight attribute level
\end{minipage} & \begin{minipage}[b]{\linewidth}\raggedleft
Human
\end{minipage} & \begin{minipage}[b]{\linewidth}\raggedleft
0.8B Base
\end{minipage} & \begin{minipage}[b]{\linewidth}\raggedleft
0.8B Pooled
\end{minipage} & \begin{minipage}[b]{\linewidth}\raggedleft
4B Base
\end{minipage} & \begin{minipage}[b]{\linewidth}\raggedleft
4B Pooled
\end{minipage} & \begin{minipage}[b]{\linewidth}\raggedleft
9B Base
\end{minipage} & \begin{minipage}[b]{\linewidth}\raggedleft
9B Pooled
\end{minipage} \\
\midrule\noalign{}
\endfirsthead
\toprule\noalign{}
\begin{minipage}[b]{\linewidth}\raggedright
Flight attribute level
\end{minipage} & \begin{minipage}[b]{\linewidth}\raggedleft
Human
\end{minipage} & \begin{minipage}[b]{\linewidth}\raggedleft
0.8B Base
\end{minipage} & \begin{minipage}[b]{\linewidth}\raggedleft
0.8B Pooled
\end{minipage} & \begin{minipage}[b]{\linewidth}\raggedleft
4B Base
\end{minipage} & \begin{minipage}[b]{\linewidth}\raggedleft
4B Pooled
\end{minipage} & \begin{minipage}[b]{\linewidth}\raggedleft
9B Base
\end{minipage} & \begin{minipage}[b]{\linewidth}\raggedleft
9B Pooled
\end{minipage} \\
\midrule\noalign{}
\endhead
\bottomrule\noalign{}
\endlastfoot
Price: \$275 & -0.085 & 0.003 & -0.070 & -0.055 & -0.088 & -0.027 &
-0.068 \\
Price: \$300 & -0.195 & -0.013 & -0.152 & -0.101 & -0.186 & -0.064 &
-0.172 \\
Price: \$325 & -0.304 & -0.020 & -0.233 & -0.164 & -0.280 & -0.127 &
-0.264 \\
Fuel: Sustainable aviation fuel & 0.069 & -0.057 & 0.010 & 0.100 & 0.079
& 0.026 & 0.049 \\
Airline: Frontier & -0.085 & 0.034 & 0.006 & -0.180 & -0.089 & -0.127 &
-0.092 \\
Airline: Southwest & -0.024 & 0.027 & 0.003 & 0.022 & 0.012 & 0.021 &
0.036 \\
Airline: United & -0.022 & 0.001 & 0.002 & -0.016 & -0.010 & -0.013 &
-0.012 \\
Baggage: Checked bag+carry-on & 0.056 & -0.043 & -0.026 & 0.087 & 0.069
& 0.086 & 0.089 \\
Baggage: Personal item only & -0.139 & -0.003 & -0.014 & -0.186 & -0.092
& -0.120 & -0.103 \\
Space: Extra legroom & 0.065 & -0.081 & -0.003 & 0.124 & 0.114 & 0.043 &
0.088 \\
Seat: Middle & -0.091 & -0.006 & -0.004 & -0.084 & -0.061 & -0.065 &
-0.063 \\
Seat: Window & 0.038 & 0.009 & 0.009 & 0.009 & 0.008 & 0.004 & 0.008 \\
\textbf{\(\mathrm{MSE}_\beta \times 10^3\)} & 0.00 & \textbf{17.75} &
\textbf{4.62} & \textbf{4.10} & \textbf{0.73} & \textbf{5.12} &
\textbf{0.93} \\
\end{longtable}
}

\emph{Note.} Human reports full-sample OLS slopes on observed respondent
choices in \texttt{Flights} (Goldfarb et al. 2026), the validation
dataset used to select each run's checkpoint. Base denotes the
off-the-shelf decision model; Pooled denotes choice training on
\texttt{Drones\ +\ Electricity}. Values back
\quartoappfigref{appfig-flights-slopes}.

\nexttablesize{\scriptsize}
{\def\LTcaptype{apptbl}
\begin{longtable}[]{@{}
  >{\raggedright\arraybackslash}p{(\linewidth - 18\tabcolsep) * \real{0.1292}}
  >{\raggedleft\arraybackslash}p{(\linewidth - 18\tabcolsep) * \real{0.0899}}
  >{\raggedleft\arraybackslash}p{(\linewidth - 18\tabcolsep) * \real{0.0899}}
  >{\raggedleft\arraybackslash}p{(\linewidth - 18\tabcolsep) * \real{0.1011}}
  >{\raggedleft\arraybackslash}p{(\linewidth - 18\tabcolsep) * \real{0.0899}}
  >{\raggedleft\arraybackslash}p{(\linewidth - 18\tabcolsep) * \real{0.0955}}
  >{\raggedleft\arraybackslash}p{(\linewidth - 18\tabcolsep) * \real{0.0899}}
  >{\raggedleft\arraybackslash}p{(\linewidth - 18\tabcolsep) * \real{0.0955}}
  >{\raggedleft\arraybackslash}p{(\linewidth - 18\tabcolsep) * \real{0.0843}}
  >{\raggedleft\arraybackslash}p{(\linewidth - 18\tabcolsep) * \real{0.0899}}@{}}
\caption{{\label{apptbl-magistro-subsets}}AI \& offshoring OLS
treatment-effect error (\(\mathrm{MSE}_\beta \times 10^3\)) by country,
arm, and attribute type}\tabularnewline
\toprule\noalign{}
\begin{minipage}[b]{\linewidth}\raggedright
Trained on
\end{minipage} & \begin{minipage}[b]{\linewidth}\raggedleft
All (32)
\end{minipage} & \begin{minipage}[b]{\linewidth}\raggedleft
US (16)
\end{minipage} & \begin{minipage}[b]{\linewidth}\raggedleft
Canada (16)
\end{minipage} & \begin{minipage}[b]{\linewidth}\raggedleft
US AI (8)
\end{minipage} & \begin{minipage}[b]{\linewidth}\raggedleft
US Off. (8)
\end{minipage} & \begin{minipage}[b]{\linewidth}\raggedleft
CA AI (8)
\end{minipage} & \begin{minipage}[b]{\linewidth}\raggedleft
CA Off. (8)
\end{minipage} & \begin{minipage}[b]{\linewidth}\raggedleft
Price (8)
\end{minipage} & \begin{minipage}[b]{\linewidth}\raggedleft
Jobs (24)
\end{minipage} \\
\midrule\noalign{}
\endfirsthead
\toprule\noalign{}
\begin{minipage}[b]{\linewidth}\raggedright
Trained on
\end{minipage} & \begin{minipage}[b]{\linewidth}\raggedleft
All (32)
\end{minipage} & \begin{minipage}[b]{\linewidth}\raggedleft
US (16)
\end{minipage} & \begin{minipage}[b]{\linewidth}\raggedleft
Canada (16)
\end{minipage} & \begin{minipage}[b]{\linewidth}\raggedleft
US AI (8)
\end{minipage} & \begin{minipage}[b]{\linewidth}\raggedleft
US Off. (8)
\end{minipage} & \begin{minipage}[b]{\linewidth}\raggedleft
CA AI (8)
\end{minipage} & \begin{minipage}[b]{\linewidth}\raggedleft
CA Off. (8)
\end{minipage} & \begin{minipage}[b]{\linewidth}\raggedleft
Price (8)
\end{minipage} & \begin{minipage}[b]{\linewidth}\raggedleft
Jobs (24)
\end{minipage} \\
\midrule\noalign{}
\endhead
\bottomrule\noalign{}
\endlastfoot
\multicolumn{10}{@{}>{\centering\arraybackslash}p{(\linewidth - 18\tabcolsep) * \real{0.9551} + 18\tabcolsep}@{}}{%
\emph{KEV-0.8B}} \\
Off-the-shelf & 33.92 & 41.48 & 26.35 & 50.62 & 32.34 & 24.20 & 28.51 &
77.60 & 19.36 \\
Electricity & 29.60 & 35.59 & 23.62 & 42.21 & 28.96 & 20.16 & 27.08 &
71.71 & 15.57 \\
Drones & 34.41 & 41.14 & 27.67 & 49.82 & 32.46 & 25.25 & 30.10 & 83.27 &
18.12 \\
Drones + Electricity & \textbf{26.61} & \textbf{32.17} & \textbf{21.06}
& \textbf{38.38} & \textbf{25.96} & \textbf{17.60} & \textbf{24.51} &
\textbf{67.37} & \textbf{13.03} \\
\multicolumn{10}{@{}>{\centering\arraybackslash}p{(\linewidth - 18\tabcolsep) * \real{0.9551} + 18\tabcolsep}@{}}{%
\emph{KEV-4B}} \\
Off-the-shelf & 99.01 & 93.76 & 104.27 & 92.30 & 95.22 & 96.88 & 111.65
& 16.09 & 126.66 \\
Electricity & 113.26 & 106.45 & 120.08 & 101.87 & 111.03 & 112.91 &
127.26 & 11.67 & 147.13 \\
Drones & 74.69 & 67.29 & 82.09 & 60.96 & 73.62 & 74.73 & 89.45 & 6.21 &
97.52 \\
Drones + Electricity & \textbf{63.98} & \textbf{56.05} & \textbf{71.90}
& \textbf{53.45} & \textbf{58.66} & \textbf{70.30} & \textbf{73.49} &
\textbf{5.66} & \textbf{83.42} \\
\multicolumn{10}{@{}>{\centering\arraybackslash}p{(\linewidth - 18\tabcolsep) * \real{0.9551} + 18\tabcolsep}@{}}{%
\emph{KEV-9B}} \\
Off-the-shelf & 82.89 & 75.34 & 90.43 & 67.71 & 82.98 & 80.55 & 100.31 &
\textbf{21.31} & 103.41 \\
Electricity & 95.68 & 83.68 & 107.69 & 71.35 & 96.01 & 98.52 & 116.86 &
37.05 & 115.23 \\
Drones & 87.42 & 79.58 & 95.26 & 68.44 & 90.72 & 86.34 & 104.18 & 37.14
& 104.18 \\
Drones + Electricity & \textbf{63.49} & \textbf{54.14} & \textbf{72.83}
& \textbf{40.89} & \textbf{67.39} & \textbf{64.66} & \textbf{81.00} &
50.56 & \textbf{67.80} \\
\end{longtable}
}

\emph{Note.} Values report \(\mathrm{MSE}_\beta \times 10^3\) on
\texttt{AI\ \&\ offshoring} (Magistro et al. 2026); lower is better.
Bold marks the lowest value per model and subset. Pooled choice training
(\texttt{Drones\ +\ Electricity}) improves every country and arm subset
across all three models, whereas single-dataset training on
\texttt{Electricity} degrades both US and Canada subsets on
\texttt{KEV-4B} and \texttt{KEV-9B}.

\end{document}